%% file: main.tex
\pdfoutput=1
\documentclass{article}

\usepackage[preprint]{neurips_2026}

\usepackage[utf8]{inputenc}
\usepackage[T1]{fontenc}
\usepackage{float}
\usepackage{array}
\usepackage[colorlinks=true,linkcolor=black,citecolor=black,urlcolor=black]{hyperref}
\usepackage{url}
\usepackage{booktabs}
\usepackage{amsmath,amssymb,amsfonts}
\usepackage{graphicx}
\usepackage{tikz}
\definecolor{preblue}{HTML}{2a78d6}
\definecolor{sftorange}{HTML}{eb6834}
\definecolor{rlgreen}{HTML}{1baf7a}
\definecolor{dpogold}{HTML}{eda100}
\usepackage{placeins}
\usepackage{microtype}
\usepackage{xcolor}

\newcommand{\relDW}{\mathrm{rel}\Delta W}
\newcommand{\astar}{\alpha^{*}}

\title{What a Weight Update Buys:\\Leverage Across Training Stages}

\author{%
  Shikhar Shiromani\\
}

\begin{document}

\maketitle

\begin{abstract}
Post-training moves a language model's weights by about a thousandth of their norm and changes its behavior
sharply; pretraining moves them by a third and buys a slow drift. We introduce \emph{leverage}, an exact measure of
what a unit of weight change buys: the Fisher quadratic form of an update, computed with one forward-mode pass per
direction, divided by the same form for random directions of the same size at the same checkpoint. Leverage orders
released Olmo-3, T\"ulu-3, Zephyr, and OLMo-2 checkpoints over four orders of magnitude: pretraining windows sit at
the random level on prose, SFT five times above it, RL-Zero and RLVR at $400$--$1{,}500$, and DPO at
$400$--$23{,}000$.
With this instrument we find that pretraining accumulates into noise: windows of 30k to 1M steps are no more
consequential on prose than random directions of their size, as a noisy quadratic model of SGD predicts. We find that
training precision changes what an update looks like: training bf16 parameters without an fp32 master copy, a common
default, freezes 36--99\% of coordinates, lowers leverage up to sevenfold, and creates an alignment between update
magnitudes and the diagonal Fisher that fp32 training does not show. We derive and verify a law for reuse: the
sign-flip null predicts the relative damage of random drop-and-rescale (DARE) at keep rate $p$ as
$(1-p)/(p\,\rho_{\text{sign}})$, within about a factor of two of measurement on released deltas wherever DARE succeeds.
Two controls make these readings reliable. Leverage is spent with displacement: every controlled SFT and DPO
trajectory with fp32 master weights and the standard losses loses it as it moves. The released level is
reproducible: two re-runs of a released DPO stage from its anchor with its published recipe land within 1\% of the
released displacement and 4\% of its leverage, where a plain recipe reads five to six times lower, and leaving one
component out, with a constant-rate control, traces the level to the length-normalized loss and its batch size of
128 together, with the data worth a factor of two.
\end{abstract}

\section{Introduction}

When a language model is fine-tuned, the change can be written as one object, the \emph{delta}: the weight
difference between the fine-tuned model and its base.
Deltas are reused: scaled and added \citep{taskarith}, sparsified and merged \citep{ties,dare}, extrapolated past
their endpoint \citep{expo}.
Some deltas tolerate this reuse and others break. The stages that produce them also differ in how far they move the
weights: a released DPO stage by $10^{-4}$--$10^{-3}$ of their norm, an SFT stage by $10^{-1}$, a stretch of
pretraining by a third.

This paper measures an exchange rate: how much behavioral change a unit of weight change buys.
We expand the KL divergence between the moved and unmoved models' next-token distributions to second order and
compute its coefficient exactly with one Jacobian--vector product. We divide it by the same quantity for random
weight changes of the same size at the same anchor (starting checkpoint).
The ratio is the \emph{leverage} $\rho$: how many times more the outputs move, in KL, than under a random change of the
same size.
The exchange rate shows three things not known before.
\emph{Pretraining accumulates into noise.} Windows of 30k to 1M pretraining steps sit at $\rho \approx 1$ on prose:
hundreds of thousands of optimizer steps add up to a direction no more consequential than a random one of its size.
A noisy quadratic model of SGD says why (\S\ref{sec:pretrain}).
\emph{A default training precision distorts what updates look like.} Training bf16 parameters with no fp32 master
copy, a common setting, freezes 36--99\% of coordinates and lowers leverage up to sevenfold. It also manufactures an
alignment, absent in fp32 training, between update magnitudes and the diagonal Fisher (\S\ref{sec:repro}).
\emph{Leverage predicts reuse.} Its sign-structure part $\rho_{\text{sign}}$ gives the relative damage of random drop-and-rescale, the operation at the
heart of DARE merging, as $(1-p)/(p\,\rho_{\text{sign}})$ at keep rate $p$, within about a factor of two of
measurement wherever DARE works (\S\ref{sec:dare}).

\begin{figure}[t]
\centering
\includegraphics[width=\linewidth]{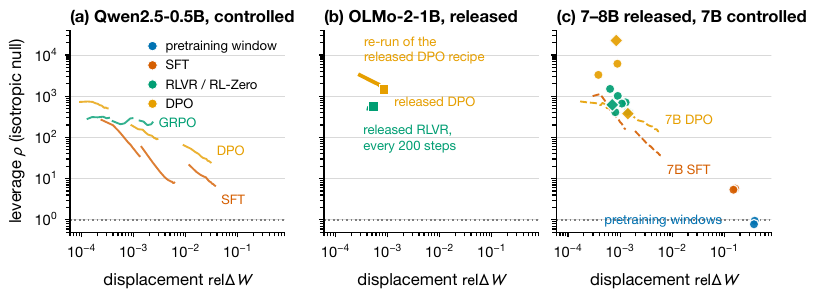}
\caption{\textbf{Leverage against displacement, measured exactly, for three model families.} Leverage $\rho$
(isotropic null; mixed corpus) against relative displacement, on the same axes in every panel; colors give the stage.
\textbf{(a)} Controlled SFT, DPO, and GRPO runs on Qwen2.5-0.5B with fp32 master weights, three learning rates each,
one seed per curve (repeats: Table~\ref{tab:r4_seeds}). \textbf{(b)} OLMo-2-1B: the released DPO delta (square), the
released RLVR checkpoints every 200 steps (line; its square is the released final), and the released DPO recipe
re-run from the released SFT checkpoint (thick line), ending on the released delta. \textbf{(c)} Released Olmo-3-7B
(circles), T\"ulu-3-8B, and Zephyr-7B (diamonds) deltas, and controlled SFT and DPO on Qwen2.5-7B (dashed, two
learning rates).}
\label{fig:main}
\end{figure}

Behind these findings sit a map of released checkpoints and two controls.
Released checkpoints span four orders of magnitude in $\rho$ (Figure~\ref{fig:main}): pretraining windows near 1,
SFT about 5, RL-Zero and RLVR $400$--$1{,}500$, DPO $400$--$23{,}000$ (\S\ref{sec:spectrum}).
The level is not a fixed property of a stage. Leverage is spent as a trajectory travels. Every controlled SFT and
DPO run with fp32 master weights and the standard losses loses it with relative displacement $\relDW$, at 0.5B and
at 7B. SFT falls as $\relDW^{-0.7}$ to $\relDW^{-1.7}$ and DPO as $\relDW^{-0.3}$ to $\relDW^{-1.2}$
(\S\ref{sec:displacement}).
Controlled training reaches the released level too. Re-running the released OLMo-2-1B DPO recipe from the released SFT
checkpoint reproduces the released delta's displacement, leverage, and sign structure. A plain recipe at the same
anchor reads five to six times lower. Leaving one component out, with a constant-rate control, finds the
length-normalized loss and the recipe's batch size of 128 carrying the level together (\S\ref{sec:repro}).

\section{Method: what a unit of weight change buys}
\label{sec:method}

Fix an anchor $\theta_0$ and a direction $\Delta$, for instance $\Delta = \theta_{\text{DPO}} - \theta_{\text{SFT}}$.
Moving along it changes the next-token distribution at each position $t$ of a fixed evaluation text, and
\[
\mathrm{KL}\big(\pi_{\theta_0 + \alpha\Delta} \,\|\, \pi_{\theta_0}\big)_t = \tfrac{\alpha^2}{2}\,
\mathrm{Var}_{p_t}\!\big(J_t\Delta\big) + O(\alpha^3),
\]
where $J_t\Delta$ is the change in the logits per unit $\alpha$ and $p_t$ is the anchor's next-token distribution.
The variance is the Fisher quadratic form $\Delta^\top F_t\Delta$, and $J_t\Delta$ is one Jacobian--vector product.
We compute $J_t\Delta$ with forward-mode automatic differentiation in float32 at every position of the corpus and
define the curvature of a direction as
\[
c(\Delta) = \frac{\tfrac12\,\mathbb{E}_t\,\mathrm{Var}_{p_t}(J_t\Delta)}{\relDW^2}, \qquad \relDW = \frac{\|\Delta\|}{\|\theta_0\|},
\]
the per-token KL cost per unit of squared relative displacement, $\relDW^2$.\footnote{$F = \mathbb{E}_t F_t$ is the Gauss--Newton part of the loss Hessian. For cross-entropy with softmax logits, it is the Hessian of the
loss in the logits pulled back through the Jacobian, which omits only the term with the logits' second derivatives.
$F$ is positive semidefinite, so $c \ge 0$. To second order every smooth $f$-divergence is a multiple of
the same form, so the choice of forward KL does not change $\rho$.}
The exact form perturbs no weights, so it has no cancellation floor.
A finite-difference estimate $\mathrm{KL}(\alpha)/(\alpha\relDW)^2$ has one: in float32 the measured KL stops falling
near $3\times10^{-8}$ nats per token, and dividing that floor by a small displacement manufactures curvature. Where
finite differences clear that floor (Appendix~\ref{app:details}), they agree with the exact form within 4\% in 649 of 659 checks (Appendix~\ref{app:r4}).

\paragraph{Leverage and its nulls.}
Leverage is $\rho = c(\Delta)/c(\text{null})$, with $c(\text{null})$ the mean over null draws made for each direction
at its own anchor. The ratio cancels the $\|\theta_0\|^2$ reparametrization factor.
The \emph{isotropic} null draws every coordinate i.i.d.\ Gaussian. Under the exact form its curvature does not depend on its
norm and averages to the mean Fisher eigenvalue, so $\rho_{\text{iso}}$ is the delta's Rayleigh quotient divided by that mean. An unsubscripted $\rho$ means $\rho_{\text{iso}}$ throughout.
The \emph{profile} null matches each tensor's norm to the delta's and gives $\rho_{\text{profile}}$.
The \emph{sign-flip} null keeps the delta's magnitudes at their positions and randomizes their signs, so its
expected form is $\sum_i F_{ii}\Delta_i^2$, the diagonal part.
Two ratios separate what an update's structure buys:
$\rho_{\text{sign}} = c(\Delta)/c(\text{sign-flip})$, the factor contributed by cross-coordinate (sign) structure,
and $c(\text{sign-flip})/c(\text{profile})$, which carries no sign information and measures how much the delta's
squared magnitudes concentrate on high-$F_{ii}$ coordinates within each tensor.
We call the second the \emph{magnitude alignment} and write it $c_{\text{sign}}/c_{\text{profile}}$.

\paragraph{Corpora, models, and precision.}
The headline corpus (the \emph{mixed} corpus below) is 51k tokens from four domains: the first 50{,}000 characters
each of web, chat, math, and code text. The appendix also reports a 766-token plain-math file and a 1{,}536-token
WikiText-2 slice.
Controlled runs train Qwen2.5-0.5B base \citep{qwen25} with SFT and DPO \citep{dpo} on UltraFeedback
\citep{ultrafeedback} at batch 8 with a constant learning rate of $5\times10^{-7}$, $5\times10^{-6}$, or $5\times10^{-5}$ for 600 steps,
measuring every 50-step checkpoint against the base.
Every controlled run except the pure-bf16 arms of \S\ref{sec:repro} keeps fp32 master weights and computes in bf16, the standard mixed-precision setting.

\section{The released spectrum}
\label{sec:spectrum}

The primary testbed is Olmo-3-7B \citep{olmo3}: two pretraining windows ending at the released base, the Instruct
and Think SFT stages, five RL-Zero runs (RLVR from the base), and the DPO and RLVR stages of the Instruct and Think
pipelines. Each is measured at the checkpoint that precedes it.
T\"ulu-3-8B \citep{tulu3} and Zephyr-7B \citep{zephyr} add post-training deltas from other pipelines, and OLMo-2-0425-1B
\citep{olmo2} adds a released DPO stage and 13 released checkpoints of its RLVR stage.

\begin{table}[t]
\centering
\small
\setlength{\tabcolsep}{3pt}
\resizebox{\linewidth}{!}{\input{tables/tab_r4_spectrum}}
\caption{\textbf{The released spectrum, measured exactly.} One row per training stage; $n$ is the number of
released directions in it, and $\relDW = \|\Delta\|/\|\theta_0\|$ is how far the stage moved the weights as a fraction
of their norm. The three leverage columns divide a direction's Fisher curvature by that of a random direction of the
same size drawn at the same anchor: an isotropic Gaussian ($\rho_{\text{iso}}$), a Gaussian with the delta's own
per-tensor norms ($\rho_{\text{profile}}$), or the delta with its signs randomized ($\rho_{\text{sign}}$); 1 means no
more consequential than random. The last column, the magnitude alignment, is the sign-flip null's curvature over the
profile null's: how much the delta's squared magnitudes sit on high-curvature coordinates. Mixed
corpus; geometric mean over the stage's directions, range in parentheses. Stages above the rule are Olmo-3-7B,
T\"ulu-3-8B, and Zephyr-7B; per-direction values, the small corpora, and the null uncertainty are in
Table~\ref{tab:r4_released} and Appendix~\ref{app:r4}.}
\label{tab:stages}
\end{table}

The two pretraining windows read 0.78 and 0.95 (Table~\ref{tab:stages}, Figure~\ref{fig:main}).
SFT reads 5.4--5.7 and the RL-Zero and RLVR directions 388--1{,}508. DPO reads 3{,}304--6{,}165 on Olmo-3, 22{,}570 on
T\"ulu-3, and 378 on Zephyr.
The per-position form is heavy-tailed; at the Olmo-3 Think anchors a few chat positions dominate the nulls, so the
Think DPO and RLVR reads carry about $\pm10\%$ uncertainty after 24 draws. Median-based reads give 5{,}238 and 544 (Appendix~\ref{app:r4}).
At 1B the order holds: the OLMo-2 DPO stage reads 1{,}452, its RLVR stage 563, and leverage stays at 480--620 along
the 13 released RLVR checkpoints (steps 200 to 2{,}600).

The nulls say where the leverage lives.
For every post-training direction, $\rho_{\text{sign}}$ is 26--129\% of $\rho_{\text{iso}}$. The sign-flip null
lowers the leverage at most fourfold (Olmo-3 Instruct DPO) and usually by less than a factor of two, so most of
it is cross-coordinate structure.
The magnitude alignment is near 1 for pretraining (0.93--1.08), SFT (1.2--1.3), RL-Zero (1.1--1.6), the
RLVR stages after DPO (1.2--2.0), and the OLMo-2 and Zephyr DPO deltas (1.27, 0.80). It is 2.2--5.3 for the Olmo-3
and T\"ulu-3 DPO deltas. Some released DPO
deltas put their mass on high-$F_{ii}$ coordinates, but high DPO leverage does not require it.
Part of that alignment can be storage: those three DPO deltas are stored in bf16, which zeroes 67--79\% of their
coordinates, and zeroing more small coordinates raises the alignment (1.26 to 1.65 for the OLMo-2 delta from 48\%
to 75\% zeros, 2.19 to 5.33 for the T\"ulu-3 DPO delta from 76\% to 92\%) while $\rho_{\text{iso}}$ stays within
10\% (Appendix~\ref{app:r4}).

\section{Leverage is spent with displacement}
\label{sec:displacement}

Released stages stop at displacements three orders of magnitude apart, so the spectrum could record how far each
stage moved rather than what kind of stage it is.
Controlled trajectories separate the two.
On Qwen2.5-0.5B with fp32 master weights and the standard losses, every SFT and DPO run loses leverage as it moves
on the mixed corpus (Figure~\ref{fig:main}a): SFT leverage falls as $\relDW^{-1.29}$, $\relDW^{-1.38}$, and
$\relDW^{-0.99}$ at the three learning rates, DPO as $\relDW^{-0.36}$, $\relDW^{-0.70}$, and $\relDW^{-0.83}$
(12 checkpoints each; seed repeats in Table~\ref{tab:r4_seeds}).
On the two small corpora the DPO run at the lowest learning rate rises instead, and every other run falls (Appendix~\ref{app:r4}).
The released OLMo-2 loss (length-normalized DPO, \S\ref{sec:repro}) is the exception on the mixed corpus: run alone,
without the rest of the released recipe, its leverage is flat or rising (slopes $+0.18 \pm 0.22$ on Qwen2.5-0.5B and $+0.32 \pm 0.19$ at the OLMo-2-1B anchor over three seeds;
Table~\ref{tab:r4_seeds}).
SFT starts lower and falls faster: at $\relDW \approx 10^{-3}$, DPO reads two to four times the SFT level.
GRPO on GSM8K from the same base, with the same precision, does not decline: at three learning rates (150 steps)
leverage stays at 200--320 on the mixed corpus (slopes $-0.27$ to $+0.25$), with $\rho_{\text{sign}}$ 150--230, like
the released OLMo-2-1B RLVR trajectory (Figure~\ref{fig:main}b).
The re-run of the released DPO recipe (\S\ref{sec:repro}) declines at the DPO slope, from $\rho = 3{,}271$--$3{,}377$ at
$\relDW = 2.9\times10^{-4}$ to 1{,}447--1{,}504 at its end ($\relDW^{-0.76}$ and $\relDW^{-0.77}$ for the two seeds).

At 0.5B the recipe knobs that set a released pipeline apart from these runs leave the curve where it is
(Figure~\ref{fig:repro}b).
Over three seeds each, DPO at learning rate $5\times10^{-6}$ reads $\rho = 158$--$188$ at $\relDW = 10^{-3}$ with
batch 8 or 128 and a constant or linearly decaying rate.
With our own trainer instead of TRL and the same plain recipe (batch 8, constant rate), DPO reads 132--180 at $10^{-3}$. Switching the loss to the length-normalized
form with $\beta = 5$ gives 90--127, switching the preference data to the released OLMo-2 mixture gives 166--175, and
starting from Qwen2.5-0.5B-Instruct instead of the base gives 143--194 (Table~\ref{tab:r4_seeds}).
Training length does not move the curve either: 10{,}000-step fp32 SFT and DPO runs at $5\times10^{-6}$ match the
600-step runs where they overlap (SFT 8.7 against 8.8, DPO 115 against 114) and continue the decline (slopes $-0.73$ and
$-0.85$; Appendix~\ref{app:traj}).

The same recipe at 7B gives the same law at a higher level (Table~\ref{tab:r5_7b}; Figure~\ref{fig:main}c).
On Qwen2.5-7B with fp32 master weights, SFT at $5\times10^{-7}$ and $5\times10^{-6}$ falls as $\relDW^{-1.46}$ and
$\relDW^{-1.19}$ and DPO as $\relDW^{-0.26}$ and $\relDW^{-0.73}$, one seed each. At matched displacement SFT reads
4--6 times its 0.5B level and DPO 1.1--2.7 times. The released 7--8B deltas stay above the controlled 7B curves: at their
own displacements the RL-Zero and RLVR deltas sit 1.2--4.5 times above the 7B SFT curve (the one curve spanning their
displacements), and the Olmo-3 and T\"ulu-3 DPO deltas 3--73 times above every controlled 7B curve that reaches them. The recipe gap of \S\ref{sec:repro} is
therefore not an artifact of the 0.5B scale.

\section{Released leverage is reproducible; pure-bf16 training distorts it}
\label{sec:repro}

\begin{figure}[t]
\centering
\includegraphics[width=\linewidth]{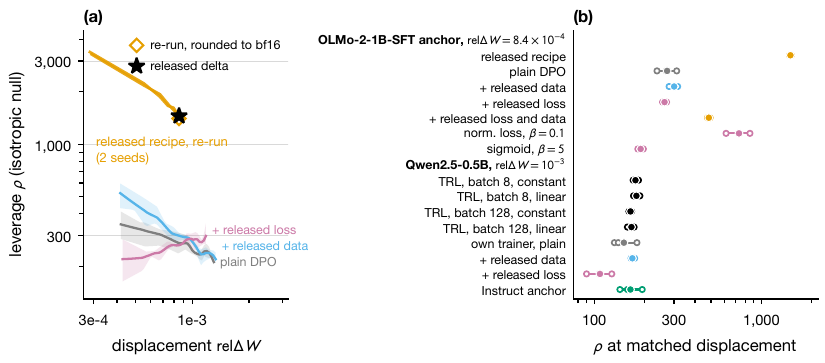}
\caption{\textbf{A released DPO stage, reproduced, and the recipe knobs under plain optimization settings.}
\textbf{(a)} The released OLMo-2-0425-1B DPO recipe re-run from the released SFT checkpoint
with fp32 master weights, two data-order seeds, measured every 296 steps. Open diamond: the final checkpoint
rounded to bf16. Star: the released DPO delta. Below: a plain DPO recipe from the same anchor and its swaps to the released data or loss (mean of three seeds,
range shaded).
\textbf{(b)} Leverage at a matched displacement for every DPO variant with seed repeats: filled, mean; open, seeds.
TRL \citep{trl} rows vary batch size and schedule at $5\times10^{-6}$; the rest use our trainer with one
component swapped, plus the loss decompositions at the OLMo-2-1B anchor (two seeds each).}
\label{fig:repro}
\end{figure}

\paragraph{The released recipe reproduces the released delta.}
If released stages wrote something controlled runs cannot, re-running a released recipe from its released anchor
would not land on the released delta.
It does (Figure~\ref{fig:repro}a), the repeat from a shared checkpoint that \citet{iclr_orient} call for.
Two re-runs with different data orders end at $\relDW = 8.594$ and $8.547\times10^{-4}$ against the released
$8.520\times10^{-4}$. They read $\rho_{\text{iso}} = 1{,}447$ and $1{,}504$ against $1{,}452$, $\rho_{\text{sign}} = 1{,}118$ and
$1{,}157$ against $1{,}012$, and a magnitude alignment of 1.27 in all three.
Rounded to bf16, the re-runs leave 47.8\% and 47.9\% of their coordinates bit-identical to the anchor; the released
delta leaves 47.9\%. Rounding moves $\rho_{\text{iso}}$ by 3\%.
The released delta's leverage lies inside the re-runs' range, and its displacement is 0.3--0.9\% below theirs. Its
$\rho_{\text{sign}}$ (1{,}012) is 9--13\% below their fp32 values and within 3\% of their bf16-rounded values (1{,}018
and 1{,}040), so that gap is storage, not training. At this anchor there is no released-only residual to explain.

\begin{figure}[t]
\centering
\begin{minipage}[c]{2.75in}\includegraphics[width=\linewidth]{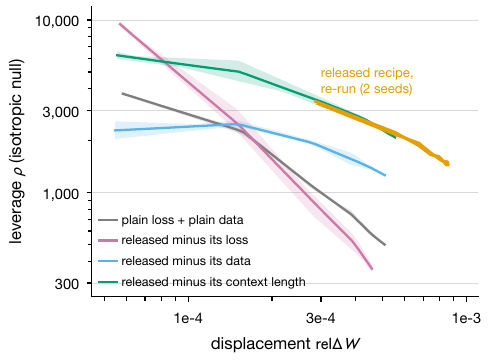}\end{minipage}\hfill
\begin{minipage}[c]{\dimexpr\linewidth-2.75in-0.18in\relax}\raggedright
\caption{\textbf{Leaving one component out of the released recipe.} Leverage against displacement at the
OLMo-2-1B SFT anchor for 600 steps of the released optimization settings (batch 128, $2.5\times10^{-6}$ with warmup
and linear decay over 2{,}955 steps, context 2{,}048). Arms: the plain loss and data, or the released recipe minus
its loss, data, or context length (three seeds; line the mean, band the range). The released re-run (two
seeds) is shown for comparison. Table~\ref{tab:loo} gives matched-displacement ratios.}
\label{fig:loo}
\end{minipage}
\end{figure}

\paragraph{What sets the level.}
The re-run reads seven to eight times the controlled 0.5B DPO runs at matched displacement. Yet at 0.5B, swapping the loss,
data, anchor, batch, or schedule moves the controlled curve by less than a factor of two (\S\ref{sec:displacement}).
The model is not the difference. A plain DPO recipe (sigmoid loss, $\beta = 0.1$, UltraFeedback, batch 8, constant
learning rate $5\times10^{-6}$) run from the same OLMo-2-1B SFT anchor reads about where it reads on Qwen2.5-0.5B
(Figure~\ref{fig:repro}b). At $\relDW = 8.4\times10^{-4}$ the released recipe reads 1{,}472--1{,}508 over two seeds and
the plain recipe 237--311 over three, a gap of 4.7--6.4 times. Under the plain settings the released loss alone gives 256--271, the released
data alone 281--315, and both together 476--493, a third of the gap on a log scale; the loss's two parts alone read
614--853 and 182--197 (Table~\ref{tab:r4_seeds}).
We then leave one component out of the released recipe, with the rest in place (batch 128, learning rate
$2.5\times10^{-6}$ with warmup and linear decay, context 2{,}048, 600 of its 2{,}955 steps; Figure~\ref{fig:loo},
Table~\ref{tab:loo}). Over three data-order seeds per arm, the arm without the released loss reads 5.5--7.3 times below the released
re-run at matched displacement. The arm without the released data reads 1.8 times below. The arm without the released context
length (768 in place of 2{,}048) matches the re-run within 4\%. The plain loss and data under the released optimization settings read 3.8--4.6 times below.
The released optimization settings therefore multiply the released loss and data by 3.1 to 5.6, depending on
the displacement at which the two settings are compared.
Seeds move each arm by at most 11\%, far less than the arms differ. The length-normalized loss and the released optimization settings together carry the level; the data is worth a
factor of two. The loss also brings the magnitude alignment of 1.23--1.31 that the released delta shows (1.27) and
that plain-loss runs lack (0.92--1.02), but only at batch 128: at batch 8 it reads 1.04--1.11, and under the plain
settings 1.00--1.06 (Table~\ref{tab:lrctl}). The minus-loss arm reverts both parts of the loss, the length
normalization and $\beta = 5$, at once and does not separate them. A constant-rate control separates the settings.
With the released loss and data at a constant $2.5\times10^{-6}$, batch 128 reads 2{,}236 (2{,}031--2{,}363) at
$4\times10^{-4}$ and 2{,}113 at $5\times10^{-4}$ over three seeds, within 20\% of the released settings (2{,}711 and
2{,}353). Batch 8 at the same rate reads 500 (363--584) and 573, where the released loss and data sit under the plain
settings (476--493; Table~\ref{tab:lrctl}). The batch size carries the factor, not the warmup, the decay, or the rate on its own. It does so
only with the released loss: at 0.5B with the plain loss, batch 128 left the level unchanged
(\S\ref{sec:displacement}).

\begin{figure}[t]
\centering
\includegraphics[width=\linewidth]{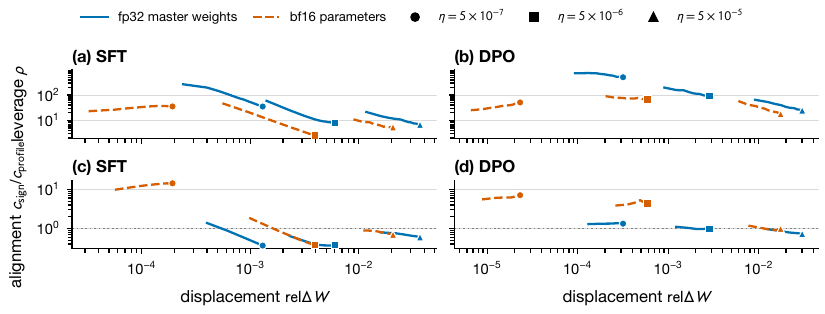}
\caption{\textbf{bf16 parameters without a master copy distort the update.} Qwen2.5-0.5B, SFT and DPO at learning
rates $5\times10^{-7}$, $5\times10^{-6}$, and $5\times10^{-5}$ (end markers), 600 steps, identical data and seed.
Blue solid: fp32 master weights. Red dashed: bf16 parameters. \textbf{(a, b)} Leverage and \textbf{(c, d)} magnitude
alignment (dotted at 1) against displacement. One seed per curve; at
$5\times10^{-6}$ three seeds differ by at most 8\% (Table~\ref{tab:r4_seeds}).}
\label{fig:precision}
\end{figure}

\paragraph{Pure-bf16 training biases update measurements.}
Loading a model in bf16 and handing it to a trainer with mixed precision enabled applies AdamW updates to bf16
parameters, and nearest rounding cancels small updates, a known effect \citep{zamirai,bf16sparsity,bf16cast}; we add
what it does to update geometry.
Adam's step is about $\eta$ per coordinate. An update smaller than half a unit in the last place rounds away, so coordinates with $|\theta|$ above $256\eta$ to
$512\eta$ never move at learning rate $\eta$: 99\% of Qwen2.5-0.5B's coordinates at $\eta = 5\times10^{-7}$, 90--91\% at $5\times10^{-6}$, and 36--46\%
at $5\times10^{-5}$.
The same recipe with fp32 master weights moves the model 1.5--14 times further at the two lower rates.
At matched displacement, pure-bf16 deltas read 1.8--7.3 times lower leverage (SFT at $\relDW = 2\times10^{-3}$: 6.7--6.8
against 31--32 over three seeds). The low-rate runs rise instead of falling
(slopes $+0.27$ and $+0.56$), and their magnitude alignment climbs to 5--14 because only
small-magnitude coordinates move (Figure~\ref{fig:precision}).
Pure-bf16 controls therefore inflate a released-versus-controlled gap and distort the alignment.

\section{Pretraining windows sit at the random level on prose}
\label{sec:pretrain}

The two Olmo-3 windows of Table~\ref{tab:stages} end at the final base, so they include mid-training, and read 0.78--1.1
on every corpus.
Windows inside pretraining stage 1 are measured with the exact form at its last checkpoint, with their own nulls at every width from 814 steps
to 1M steps (Table~\ref{tab:r4_stage1}). From 30k steps on ($\relDW$ 0.10--0.92) they read the random level on prose:
$\rho_{\text{iso}} = 0.6$--$1.5$ on plain-math, WikiText, web, and chat text. On math and code text the same windows read 0.6--25. The pooled mixed read is 0.8--12 while its median
stays at 0.7--1.0, because the top 1\% of positions carry 21--72\% of a window's form, against 20\% for the nulls
(Appendix~\ref{app:pythia}). Narrower windows read higher on every text. At 814 steps they read 3.9--15 on prose and 750 on math, then fall as
width$^{-0.5}$ to width$^{-0.7}$ on prose and width$^{-1.7}$ on math to the plateau by about 30k steps. Finite
differences at $\alpha = 0.3$, where the quadratic term swamps those positions, read the narrow windows 1.1--5 times
lower on prose (Appendix~\ref{app:pythia}). Pythia-1.4B
\citep{pythia}, measured the same way backward from step 143k, reads 5.0, 3.0, 1.9, 1.1, 0.58, and 0.65 on the mixed
corpus at 1k, 2k, 4k, 8k, 16k, and 43k steps. These reads sit 10--46\% below finite differences, and the gap widens
with width. No positions dominate (top-1\% share 0.11--0.16; Table~\ref{tab:r5_pythia}). Both lineages reach the prose plateau ten to thirty thousand steps
before the anchor, and $\rho_{\text{sign}}$ falls from 616 (Olmo) and 14 (Pythia) at the narrowest width to 0.8--4.2 on
the plateau, except 19 at the 30k Olmo width, where the pooled read is 12.

\paragraph{A noisy quadratic model of the plateau.}
Near a minimum where the loss Hessian is the Fisher and minibatch gradient noise has covariance $F/B$ at batch size $B$,
SGD with step $\eta$ has a stationary weight covariance that solves $F\Sigma + \Sigma F = (\eta/B)F$ \citep{mandt,nqm,liuziyin}.
The solution is $\Sigma = (\eta/2B)\,I$, so late-training fluctuations are isotropic in weight space.
A window of width $w$ in that regime has $\mathbb{E}[\Delta\Delta^\top] = (\eta/B)(I - e^{-\eta w F})$, and its
leverage is the mean Fisher eigenvalue weighted by how far each mode has equilibrated, divided by the unweighted mean.
When every mode has equilibrated, $\rho = 1$ whatever $\eta$ and $B$ are. For a single step the noise
concentrates in the top eigenspace, as \citet{gurari2018} observed for gradients, and the ratio of expected forms
tends to $d\sum_k\lambda_k^2/(\sum_k\lambda_k)^2$ over the $d$ modes (the expected Rayleigh quotient sits below it,
because the top few modes dominate).
For a power-law spectrum $\lambda_k \propto k^{-a}$, the transition between these two limits gives
$\rho \propto w^{1/a-2}$ while modes are still equilibrating. The measured $w^{-0.5}$ to $w^{-0.7}$ on prose
corresponds to $a = 0.67$--$0.75$, an exponent we do not measure; a finite spectrum of $2\times10^6$ modes flattens
the prediction to $w^{-0.41}$ to $w^{-0.54}$ (Appendix~\ref{app:nqm}).
The model's displacement grows diffusively, $\|\Delta\| \propto w^{0.5}$, against the measured $w^{0.35}$--$w^{0.42}$;
the same $2\times10^6$ modes give $w^{0.42}$--$w^{0.44}$ from the cutoff alone.
The model describes the decline of $\rho$ with one free exponent; it does not explain the sub-diffusion.
Because $\Sigma \propto I$, it predicts $\rho = 1$ on any text; the math-text excess is a drift the stationary model
leaves out. It also holds $F$ and $\eta$ fixed, which fails over tens of thousands of steps. A backward Pythia window anchored at its end reads 1.5--2.0 times above a forward window of the same width anchored
at its start, and the one delta measured at both anchors reads 3.5--5.6 times higher at the end.
Olmo displacement turns linear in width beyond 400k steps, where the window reaches the early high-rate part of the run (Appendix~\ref{app:pythia}).
In the model the noise isotropy sets the plateau; batch and step size do not move it. A diagonal preconditioner,
as in Adam, tilts the stationary covariance toward flat coordinates ($\Sigma_{ii} \propto F_{ii}^{-1/2}$) and puts
the plateau at or below 1 for any Fisher. The $w^{-1/2}$ transition is then equilibration across the off-diagonal
spectrum, which the sign-flip null supports: the narrow-window excess sits in cross-coordinate structure
($\rho_{\text{sign}} = 616$ and 14 at the narrowest Olmo and Pythia widths; Appendix~\ref{app:nqm}).

\section{Leverage predicts what survives drop-and-rescale}
\label{sec:dare}

\begin{figure}[t]
\centering
\begin{minipage}[c]{2.85in}\includegraphics[width=\linewidth]{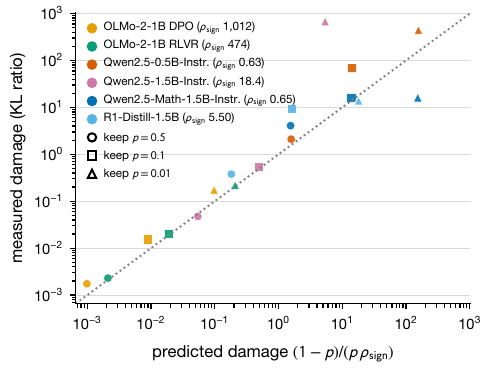}\end{minipage}\hfill
\begin{minipage}[c]{\dimexpr\linewidth-2.85in-0.18in\relax}\raggedright
\caption{\textbf{The sign-flip null predicts DARE damage until DARE breaks.} Measured damage, the KL between
the DARE'd and the unpruned model relative to the delta's own KL, both at full scale (mixed corpus, mean of two masks), against
$(1-p)/(p\,\rho_{\text{sign}})$ for six released deltas at keep rates $p = 0.5$ (circles), 0.1 (squares), and 0.01
(triangles). Dotted: equality.}
\label{fig:dare}
\end{minipage}
\end{figure}

DARE \citep{dare} keeps each coordinate of a delta with probability $p$ and rescales the kept ones by $1/p$. Its
authors find that it works when delta entries are small. The quadratic form says when.
For a mask $m$ with independent Bernoulli($p$) entries,
\begin{equation}
\mathbb{E}_m\big[(m\odot\Delta/p)^\top F\,(m\odot\Delta/p)\big] = \Delta^\top F\Delta + \frac{1-p}{p}\sum_i F_{ii}\Delta_i^2,
\label{eq:dare}
\end{equation}
and the sum in the second term is the sign-flip null's form.
Two analyses precede this one. A first-order expansion shows that DARE leaves the loss unchanged to first order
\citep{deltaedit}, which cannot say when DARE fails. A per-layer bound ties failure to the rescaling factor and to
the delta's contribution to each activation \citep{darex}; its variance factor $q/(1-q)$ ($q$ their drop rate) is the
Bernoulli factor of Eq.~\eqref{eq:dare} placed on hidden states rather than in the Fisher metric.
Eq.~\eqref{eq:dare} is second order, concerns the whole output distribution, and becomes a single number once
divided by the delta's own form.
We measure the damage from the fine-tuned endpoint. The pruning residual $r = \mathrm{DARE}_p(\Delta) - \Delta = (m/p - \mathbf{1})\odot\Delta$ has
zero mean and covariance $\tfrac{1-p}{p}\,\mathrm{diag}(\Delta_i^2)$, so to second order
$\mathbb{E}_m\,\mathrm{KL}(\theta_0 + \mathrm{DARE}_p(\Delta)\,\|\,\theta_0 + \Delta) = \tfrac12\tfrac{1-p}{p}\sum_i
F_{ii}(\theta_0 + \Delta)\,\Delta_i^2$, the sign-flip form at the endpoint.
Dividing by the delta's own KL, $\tfrac12\Delta^\top F(\theta_0)\Delta$, gives the relative damage. When the diagonal
Fisher is the same at anchor and endpoint, the prediction is $(1-p)/(p\,\rho_{\text{sign}})$; otherwise it carries an
extra factor, the endpoint-to-anchor ratio of the sign-flip null's curvature.
We measured the damage at full scale, the KL itself rather than its quadratic form, on six released deltas whose
$\rho_{\text{sign}}$ spans 0.6 to 1{,}012, at keep rates $p = 0.5$, 0.1, and 0.01 (Figure~\ref{fig:dare}): the KL
between the DARE'd and the unpruned model divided by the KL of the delta (mixed corpus, two masks, which agree
within 1--21\%). With the quadratic form as denominator, the four
small deltas move by at most 6\% and the two largest deltas (by displacement) rise 1.2--2.1 times, because their full-scale KL exceeds the form.
Wherever the predicted damage is below about 0.5, the measurement tracks it within a factor of 2.1.
OLMo-2-1B RLVR ($\rho_{\text{sign}} = 474$) loses 0.2\%, 2.0\%, and 22\% of its own KL at the three keep rates,
against predicted 0.2\%, 1.9\%, and 21\%. Qwen2.5-1.5B-Instruct ($\rho_{\text{sign}} = 18.4$) loses 4.8\% and 53\%
at $p = 0.5$ and 0.1, against 5.4\% and 49\%.
The OLMo-2-1B DPO delta reads a constant 1.75 times its prediction at every $p$, and the isotropic null's
curvature at its endpoint is 1.68 times that at its anchor, the scaling the expansion predicts if the diagonal
Fisher grows in step with the mean eigenvalue.
R1-Distill, a delta of 12\% of the weight norm, reads 2.1 times its prediction already at $p = 0.5$.
Once the prediction passes about 1, damage no longer follows the quadratic and DARE has failed; SFT-like deltas
with $\rho_{\text{sign}} \lesssim 6$ reach that point by $p = 0.1$. For the two largest deltas at $p = 0.01$ the
damage falls below the prediction.
The sign-flip null therefore sets how much of a delta can be dropped: a keep rate of about $1/(1 +
\epsilon\rho_{\text{sign}})$ for a tolerated relative damage $\epsilon$.

Extrapolation is another reuse the introduction named; our result there is inconclusive. We preregistered two
predictors of whether a delta keeps its gain at $1.5\times$, the quadratic radius $\astar$ of Appendix~\ref{app:astar}
and the displacement. The first scored 5 of 8 scorable deltas (of 17) on GSM8K; the second 3 of 3 (of 12) on GSM8K
and 1 of 5 (of 21) on MATH-500. The confidence intervals
exclude little (Appendix~\ref{app:astar}). The raw few-shot prompts left 10 of 21 MATH-500 endpoints below their
anchors before any extension, and extension often broke the output format before the arithmetic. \citet{iclr_traj}
find gains that peak 0.05 of the delta beyond the endpoint and fade by 0.5, where our tests sit.

\section{Related work}
\label{sec:related}

The instrument's parts are established: directional sharpness as an optimizer diagnostic
\citep{panli2023,wang2026muon,kalra2026}, random-perturbation baselines and fixed-direction landscape probes
\citep{keskar,goodfellow,lilandscape,andriushchenko2023}, the Fisher form as a sensitivity measure \citep{ewc}, and
the KL quadratic's radius in trust-region methods \citep{trpo,martens}. We aim them at what kind of training produced
an update, at matched displacement, with the form computed exactly.
Four concurrent submissions compare trained directions with random ones. \citet{iclr_orient} rotate RLVR
updates, norm and spectrum preserved, and read the effect on entropy. \citet{iclr_traj} compare a fine-tuning
displacement with isotropic search, and its loss-Hessian Rayleigh quotient with random directions.
\citet{iclr_perturb} probe loss curvature along a sparse rollback direction and count-matched random directions.
\citet{iclr_onebit} draw sign-shuffle nulls for sparse behavioral edits. The Rayleigh quotient and the rollback probe
are finite-difference loss Hessians on a batch or two; one spans 0.2--372 across models, the other changes sign along
a trajectory. None of the four computes the Fisher form exactly, draws nulls for that form at the delta's anchor, or
reads the sign-flip null as an update's diagonal-Fisher share.
Recent work characterizes RLVR's footprint at the token level \citep{sparsecritical} and in weight space through
sparsity and spectral statistics \citep{dposparsity,sparsitycurse,manytickets,pathnottaken,offlinegeom,blocklevel}. Our
metric is the function-space Fisher, our nulls keep the delta's structure, and we control displacement.
\citet{fishersparsity} explain RLVR's sparsity through the diagonal Fisher, the quantity our magnitude alignment
registers. \citet{bf16sparsity} trace much of the sparsity to bf16 rounding, and \S\ref{sec:repro} shows the same on SFT and DPO.
\citet{dco} use the same quadratic as a drift budget, task gain per unit of behavioral radius; $\rho$ is the
complementary quantity, behavioral change per unit of weight motion.
\citet{picozzi} applies the same output metric to activation interventions, \citet{ruscio2026} contrast pretraining
and alignment deltas against direction-matched random controls, and \citet{esvsgrpo} report that GRPO moves weights
two orders of magnitude less than evolution strategies at the same accuracy.
On reuse, task arithmetic \citep{taskarith} adds deltas, TIES-Merging \citep{ties} resolves their signs, BitDelta
\citep{bitdelta} binarizes them, and DARE \citep{dare} sparsifies them. \citet{iclr_onebit} find that randomizing a
behavioral edit's signs removes its effect while binarization keeps it, the behavioral face of $\rho_{\text{sign}}
\gg 1$. Section~\ref{sec:dare} gives DARE's tolerance in closed form.
ExPO \citep{expo} and \citet{overthinking} extrapolate and amplify deltas. \citet{rlvrlinear} report near-linear RLVR
dynamics, consistent with the flat leverage of our RLVR runs.
\emph{RL's Razor} \citep{razor} predicts forgetting from total KL, whereas leverage is KL per unit of weight motion.
\citet{kishino} fit KL-versus-distance exponents within runs.

\section{Limitations and conclusion}
\label{sec:limits}

\paragraph{Limitations.}
Controlled comparisons use two or three seeds and the reproduction two. Seed ranges of leverage at matched displacement are
at most 9\% (TRL) and 48\% (our trainer) of the mean, below every effect we report \citep[Tables~\ref{tab:r4_seeds} and~\ref{tab:lrctl};
run-to-run noise per][]{runtorun}. Controlled runs are at 0.5B and 1B with seed repeats and at 7B with one seed per arm; our GRPO runs are short
(150 steps); the reproduction covers one released DPO stage at 1B. The leave-one-out covers 600 of the recipe's 2{,}955 steps; its
constant-rate control separates schedule from batch size, and tests a second learning rate only at batch 8, not at batch 128.
The extrapolation tests rest on finite-difference reads and few scorable deltas. Released deltas are stored in bf16, which
leaves 26--79\% of the 7--8B RL and DPO deltas' coordinates at their anchors; at 48\% zeros storage moved leverage 3\%,
alignment under 2\%, and $\rho_{\text{sign}}$ 9--10\%, but further zeroing raises the alignment and lowers
$\rho_{\text{sign}}$, so the sparsest stored deltas' alignments are upper bounds (Appendix~\ref{app:r4}).

\paragraph{Conclusion.}
Leverage, the Fisher cost of an update's direction relative to random directions of its size, separates training
stages by four orders of magnitude and falls as an SFT or DPO update travels; re-running a released recipe from its
anchor reproduces it. The recipe sets the level through its loss and its batch size together, and training precision
moves it as much as the recipe does. The sign-structure part of leverage predicts what survives random
sparsification. A noisy quadratic model predicts the approach to 1 on prose in pretraining.
Anyone dividing a small divergence by a small displacement should compute the form exactly, draw the null at the
delta's anchor, and train with fp32 master weights.

\paragraph{Impact statement.}
We measure weight updates of public language models with public data and create no new capability. Showing which
updates are cheap to sparsify bears on the reuse and misuse of open checkpoints.

\newpage
\begingroup\raggedright

\endgroup

\appendix

\section{The exact instrument, controlled runs, and reproduction}
\label{app:r4}

\paragraph{Computation.}
For a direction $v$ at anchor $\theta_0$ and each window of at most 2{,}048 tokens, one call to
\texttt{torch.func.jvp} through \texttt{functional\_call} returns the anchor's logits $z_t$ and their directional
derivative $J_tv$. With $p_t = \mathrm{softmax}(z_t)$, the per-position quadratic coefficient is
$q_t = \tfrac12(\sum_y p_{t,y}(J_tv)_y^2 - (\sum_y p_{t,y}(J_tv)_y)^2)$, and $c(v) = \mathbb{E}_t q_t / \relDW^2$.
Everything runs in float32 with TF32 disabled and eager attention. One pass costs about three forward passes. The
mean, median, and top-1\% and top-10\% shares of $q_t$ are stored per corpus.
Each corpus file is tokenized as one string with the measured anchor's own tokenizer, which adds its default special tokens once per string. The Qwen2.5, Olmo-3, and OLMo-2 tokenizers add none. The Llama-3.1 and Mistral anchors' tokenizers prepend a BOS token, so it appears only at the start of
the first window of each file. The string is cut into contiguous,
non-overlapping windows of at most 2{,}048 tokens. Each window is a separate forward pass with no cross-window context. A
trailing window is kept when it has at least two tokens. $\mathbb{E}_t$ runs over every predicted position,
so windows and domains weigh by their token counts. Tied embeddings are one parameter: counted once in
$\|\theta_0\|^2$, $\|\Delta\|^2$, and the parameter count $d$, and carried by one tangent tensor that feeds both the
embedding lookup and the logit projection.
Isotropic draws are i.i.d.\ $\mathcal{N}(0,1)$ in every tensor with no per-tensor normalization, so the direction is
uniform on the sphere and $\mathbb{E}[c] = \tfrac12\|\theta_0\|^2\,\mathrm{tr}F/d$. The exact form is scale-free, so no rescaling to $\|\Delta\|$ is needed. In this campaign we apply finite differences
to the measured deltas only and measure every null draw with the exact form. Profile draws rescale each tensor's Gaussian to the delta's tensor norm, where tensors are the entries of the
model's \texttt{named\_parameters} (unfused projections and norms as their own tensors, tied embeddings once). Sign-flip
draws multiply the delta by independent random signs. The measurement code is \texttt{experiments/round4/r4\_measure.py}; the
controlled runs and the reproduction use \texttt{r4.py} and \texttt{r4\_dpo.py} beside it. Released endpoints whose configuration \texttt{transformers} does not
recognize (Olmo-3-7B-RL-Zero-Mix ships an \texttt{olmo2-retrofit} configuration) are read from their safetensors by
parameter name, and every tensor matched.

\paragraph{Agreement with finite differences.}
Every checkpoint of the 600- and 300-step runs of Table~\ref{tab:r4_controlled} and their seed repeats was also measured by finite differences at $\alpha \in \{0.01, 0.03, 0.1, 0.3\}$ on
plain-math and WikiText. Where the floor and ratio tests of the earlier campaign's certification (Appendix~\ref{app:details}) pass ($\mathrm{KL}(0.1) > 10^{-6}$ and
$\mathrm{KL}(0.1)/\mathrm{KL}(0.03) \in [7, 17]$), the finite-difference curvature at $\alpha = 0.1$ is within 4\%
of the exact value in 649 of 659 checks below $\relDW = 10^{-2}$ and within 15\% in all of them. Beyond
$\relDW \approx 10^{-2}$ the finite-difference read is 0.71--1.02 times the exact value, because $\alpha = 0.1$ is no longer quadratic
there. Where the screen fails, the finite-difference value exceeds the exact one by up to $72\times$ (pure-bf16 DPO
at learning rate $5\times10^{-7}$). There the read is the float32 arithmetic floor, near $3\times10^{-8}$ nats per token, divided by a
small displacement. The exact form needs no screen and no certified scale.

\paragraph{Null stability.}
$c(\text{null})$ is the mean over draws. The released fan and the pretraining windows use 6 isotropic draws per anchor and 3 profile and 3 sign-flip draws per direction; the controlled runs and the OLMo-2-1B released deltas use 8, 2, and 2; the 7B runs use 6, 2, and 2. The isotropic curvature of Qwen2.5-0.5B on the mixed corpus is $97.7 \pm 0.47$ (standard error over 8 draws).
On the mixed corpus the draw-to-draw standard deviation is 1--3\% of the mean at 11 of 16 anchors and 6--17\% at the
Olmo-3 Instruct-SFT, Instruct-DPO, and stage-1 anchors. At the Olmo-3 Think anchors a few chat positions dominate the
nulls. The top 1\% of positions carry 30--37\% of the sum, and draws scatter by 56\% and 36\%. We therefore use 24 isotropic
draws there (standard error 11\% and 7\%) and 12 profile and 12 sign-flip draws for the Think DPO and RLVR deltas.
Their sign-flip draws still scatter by 75\% and 45\%, so the Think magnitude alignments (5.3 and 2.0) are uncertain
by about a fifth. Median-based reads, which the dominant positions do not move, give $\rho_{\text{iso}} = 5{,}238$ and 544.
On the two small corpora draws scatter by 3--21\%.

\begin{table}[H]
\centering
\scriptsize
\setlength{\tabcolsep}{3pt}
\input{tables/tab_r4_released_full}
\caption{\textbf{Every released direction, exact form.} Columns as in Table~\ref{tab:stages}, one row per direction. Leverage against isotropic nulls on the three corpora, and
against profile and sign-flip nulls on the mixed corpus (6 isotropic draws per anchor and 3 profile and 3 sign-flip draws per direction; 24, 12, and 12 at the Think anchors; 8, 2, and 2 for the OLMo-2-1B rows); ``unchanged'' is the fraction of coordinates bit-identical
to the anchor (bf16 storage). ``FD mixed'' is the earlier finite-difference read at the certified common scale
(Appendix~\ref{app:details}); \texttt{dpo\_think} has no common certified scale there and the OLMo-2-1B deltas were not in that campaign.}
\label{tab:r4_released}
\end{table}

\begin{table}[H]
\centering
\scriptsize
\setlength{\tabcolsep}{2.5pt}
\input{tables/tab_r4_controlled}
\caption{\textbf{Every controlled run on Qwen2.5-0.5B} (UltraFeedback, 600 steps for batch 8 and 300 for batch 128,
seed 0; the GRPO rows train on GSM8K for 150 steps and the last two rows for 10{,}000 steps). ``unchanged'' is the fraction of coordinates bit-identical to the base at the last checkpoint (seed 0; the other
seeds of the $5\times10^{-5}$ pure-bf16 runs reach 0.46). Slopes are
ordinary least squares of $\log c$ on $\log\relDW$ over all checkpoints of a run (12; 6 for GRPO, 40 for the 10{,}000-step runs); the last column ranges over
those with nulls (6; 3 and 10).}
\label{tab:r4_controlled}
\end{table}

\paragraph{The reproduction.}
The released OLMo-2-0425-1B DPO recipe (\texttt{docs/olmo2.md} in open-instruct, \citealp{openinstruct}) is length-normalized DPO with $\beta = 5$,
learning rate $2.5\times10^{-6}$, effective batch 128, linear schedule with 10\% warmup over one epoch of
\texttt{allenai/olmo-2-0425-1b-preference-mix} (378{,}301 pairs, 2{,}955 steps), maximum length 2{,}048, weight
decay 0, AdamW $(0.9, 0.999, 10^{-8})$, BOS prepended, the tokenizer's chat template, no gradient clipping (we
assume open-instruct's default). We ran it from \texttt{allenai/OLMo-2-0425-1B-SFT} with fp32 master weights and
bf16 autocast on two H100s with data parallelism (2 ranks $\times$ 4 pairs $\times$ 16 accumulation steps), a frozen
bf16 reference model, and data order shuffled with seed 111 or 222; each run took 8.6 hours. Final training accuracy and margin were 0.68 and 0.54 for seed 111 and 0.63 and 0.65 for seed 222. The final checkpoint was measured in fp32 and after rounding every parameter to bf16.

The same trainer ran the released recipe and the single-knob variants of \S\ref{sec:displacement} and
\S\ref{sec:repro}; each variant ran 600 steps at batch 8, constant learning rate $5\times10^{-6}$, maximum length
768, no gradient clipping, on UltraFeedback-binarized \texttt{train\_prefs} (or the released mixture) shuffled with
the run's seed, from the anchor shown (Table~\ref{tab:r4_repro}). Every run of this trainer renders prompt and
response through the tokenizer's chat template with BOS prepended and masks the prompt, so the plain and released arms
differ only in the knobs named. Both losses are $-\log\sigma\big(\beta[(\ell_\theta(y_w) - \ell_{\text{ref}}(y_w)) -
(\ell_\theta(y_l) - \ell_{\text{ref}}(y_l))]\big)$ against a frozen reference, with $\ell$ the summed response
log-probability (\texttt{sigmoid}) or its per-token mean (\texttt{dpo\_norm}).

\begin{table}[H]
\centering
\scriptsize
\setlength{\tabcolsep}{3pt}
\resizebox{\linewidth}{!}{\input{tables/tab_r4_repro_family}}
\caption{\textbf{Runs of the reproduction trainer.} The released OLMo-2-1B recipe and single-knob variants of a plain
DPO recipe at three anchors; leverage at the first and last checkpoint, slope of $\log c$ on $\log\relDW$, and the
range of the magnitude alignment over checkpoints with nulls. One seed each; Table~\ref{tab:r4_seeds} gives the
repeats.}
\label{tab:r4_repro}
\end{table}

\begin{table}[H]
\centering
\scriptsize
\setlength{\tabcolsep}{3pt}
\input{tables/tab_r4_seeds}
\caption{\textbf{Seed repeats.} Leverage at a reference displacement (mixed corpus; mean and range over seeds) and the
slope of $\log c$ on $\log\relDW$ (mean $\pm$ s.d.), for every controlled comparison in the body. Seeds change the data order; the released-recipe re-run uses data-order seeds 111 and 222. The pure-bf16 DPO run at
$5\times10^{-6}$ stops at $\relDW = 5.9\times10^{-4}$, so its reference displacement is $5\times10^{-4}$.}
\label{tab:r4_seeds}
\end{table}

\paragraph{Leave-one-out at the released settings.}
Four arms of the reproduction trainer start from the OLMo-2-1B SFT anchor with three data-order seeds each (111, 222, 333).
All keep the released optimization settings (batch
128 as 4 pairs $\times$ 32 accumulation steps, learning rate $2.5\times10^{-6}$, 10\% warmup and linear decay
scheduled over 2{,}955 steps, no clipping) and context 2{,}048 unless noted, stop at step 600, and are measured every 100 steps. The arms are the plain loss and data (sigmoid, $\beta = 0.1$, UltraFeedback), the released recipe minus its loss
(sigmoid, $\beta = 0.1$, OLMo-2 mixture), minus its data (length-normalized, $\beta = 5$, UltraFeedback), and minus its
context length (length-normalized, $\beta = 5$, OLMo-2 mixture, context 768). Table~\ref{tab:loo} compares each at
$\relDW = 4\times10^{-4}$ and $5\times10^{-4}$ with the released re-run interpolated at the same displacement.
Three further variants under the plain settings (batch 8, constant $5\times10^{-6}$, context 768, 600 steps, two seeds
each) isolate the loss knob: the released loss with the released data, the length-normalized loss at $\beta = 0.1$, and
the sigmoid loss at $\beta = 5$ (Table~\ref{tab:r4_seeds}).

\begin{table}[H]
\centering
\scriptsize
\setlength{\tabcolsep}{3pt}
\resizebox{\linewidth}{!}{\input{tables/tab_r5_loo}}
\caption{\textbf{Leave-one-out at the released settings}, OLMo-2-1B SFT anchor, three data-order seeds per arm:
leverage at two matched displacements (seed mean and range) against the released re-run (mean of two seeds,
interpolated), the slope of $\log c$ on $\log\relDW$ over the six checkpoints (mean $\pm$ s.d.), and the range over
seeds of the last checkpoint's magnitude alignment and $\rho_{\text{sign}}$. One seed of the minus-loss arm met a
gradient spike at step 558 (gradient norm 727 against a median of 32 and no other step above 163; the recipe does not clip). Its
step-600 checkpoint then carries 93\% of its form in 1\% of positions. We exclude that checkpoint; its median-based
read, 228, continues the arm's decline. The $5\times10^{-4}$ cell of that arm therefore averages the two seeds whose
clean trajectories reach it. The upper end of its $\rho_{\text{sign}}$ range (524) is the spiking seed's step-400
checkpoint, its last with nulls.}
\label{tab:loo}
\end{table}

\begin{table}[H]
\centering
\scriptsize
\setlength{\tabcolsep}{3pt}
\resizebox{\linewidth}{!}{\input{tables/tab_r5_lrctl}}
\caption{\textbf{Constant-rate control for the leave-one-out.} The released loss and data at the OLMo-2-1B SFT anchor
with a constant learning rate of $2.5\times10^{-6}$, no warmup or decay, at batch 128 (600 steps) and batch 8 (800
steps), three data-order seeds each; columns as in Table~\ref{tab:loo}.}
\label{tab:lrctl}
\end{table}

\begin{table}[H]
\centering
\scriptsize
\setlength{\tabcolsep}{4pt}
\input{tables/tab_r5_7b}
\caption{\textbf{The 0.5B recipe at 7B.} SFT and DPO on Qwen2.5-7B with fp32 master weights, batch 8, constant learning
rate, 600 steps, one seed each: displacement range, leverage at three displacements where the run covers them, slope of
$\log c$ on $\log\relDW$, the range of the magnitude alignment over checkpoints with nulls, and the fraction of
coordinates unchanged at the last checkpoint (unused vocabulary rows).}
\label{tab:r5_7b}
\end{table}

\paragraph{GRPO and long runs.}
The GRPO arms train Qwen2.5-0.5B with TRL on the 7{,}473 GSM8K training problems (prompt ``Question: \ldots
Answer:'') with fp32 master weights. Each prompt gets four generations of 200 completion tokens; the KL coefficient is $\beta = 0$, and the reward is 1 when the last number
in the completion equals the gold answer. Batch is 8, and each run lasts 150 steps, measured every 25. The
10{,}000-step SFT and DPO runs at $5\times10^{-6}$ cover 1.3 epochs of UltraFeedback-binarized and are measured every
250 steps, with nulls on every fourth checkpoint. Both appear in Table~\ref{tab:r4_controlled}.

\paragraph{Training precision.}
A common setup loads the model in bf16 (\texttt{dtype=torch.bfloat16}) and hands it to a trainer with
\texttt{bf16=True}. The optimizer then updates bf16 parameters directly, with no fp32 master copy. Adam's step is
about the learning rate per coordinate, and a bf16 weight moves only when the step exceeds half a unit in its last
place, between $|\theta|\,2^{-9}$ and $|\theta|\,2^{-8}$. Coordinates with $|\theta|$ above $256\eta$ to $512\eta$
therefore never move. \citet{bf16sparsity} report the per-step counterpart on the same model. At $\eta = 3\times10^{-6}$ with fp32 master weights, 96\% of its weights sit
above the rounding threshold, so the bf16 cast hides an update
that the master copy keeps. Our fractions are cumulative over training with no master copy, where the update is lost.
A 40-step check on Qwen2.5-0.5B left 93\% of coordinates bit-identical at $\eta = 5\times10^{-6}$
(none with $|\theta_0| > 3\times10^{-3}$ moved, a threshold slightly above $512\eta = 2.6\times10^{-3}$ because Adam's step can exceed
$\eta$ for a few steps, up to $(1-\beta_1)/\sqrt{1-\beta_2} \approx 3\eta$) and 99\% at $5\times10^{-7}$, with a half and a tenth of the fp32-master
displacement. Every controlled run of the earlier finite-difference campaign (Appendix~\ref{app:traj}) used this
setting. Their displacements, slopes, and magnitude alignments carry the bias of \S\ref{sec:repro}, and the
exact-form runs with fp32 master weights replace them in the body.

\paragraph{Storage truncation.}
A released delta is the difference of two bf16-stored checkpoints, so every coordinate whose update was below about
half a unit in the last place (ulp) of the anchor value is zero: 48\% of the coordinates at 1B and 26--79\% for the 7--8B
RL and DPO deltas. To ask what further zeroing does, Table~\ref{tab:r5_trunc} zeroes every coordinate below $t$
units in the last place for $t = 1, 2, 4$ (and 8 at 1B) and re-measures the delta with its own nulls. At $t = 1$
almost nothing changes, because the stored coordinates are whole multiples of the ulp. Beyond that, $\rho_{\text{iso}}$ holds
within 10\% for the DPO deltas up to about 90\% zeros and falls after; the T\"ulu-3 RLVR delta halves already at
93\%. $\rho_{\text{sign}}$ falls from $t = 2$ on. The magnitude alignment rises steeply: 1.26 to 3.17 at 1B,
2.19 to 10.1 for the T\"ulu-3 DPO delta, and 3.55 to 12.1 for Olmo-3 Instruct DPO. Zeroing small coordinates removes the
cross-coordinate share of the form, and the norm with it, faster than the diagonal share that the sign-flip null
measures, which is what inflates the ratio. The alignments of the sparsest stored deltas are therefore upper bounds on their fp32 training values;
$\rho_{\text{iso}}$ at the stored level is little affected for the DPO deltas, whose ladders hold to about 90\% zeros.

\begin{table}[H]
\centering
\scriptsize
\setlength{\tabcolsep}{4pt}
\input{tables/tab_r5_trunc}
\caption{\textbf{Truncation ladder.} Four released deltas as stored and with every coordinate below $t$ bf16 units in
the last place of the anchor value zeroed; each row is a full measurement with 6 isotropic, 3 profile, and 3 sign-flip
draws on the mixed corpus; Table~\ref{tab:r4_released} uses 8, 2, and 2 draws for the OLMo-2-1B delta, so its values differ from the as-stored row here in the third digit.}
\label{tab:r5_trunc}
\end{table}

\section{Measurement details}
\label{app:details}

\paragraph{What the instrument cannot see.} $F$ is positive semidefinite, so directions of negative loss curvature are
invisible to it. $c$ is only the Gauss--Newton part of directional sharpness, and it also
carries a $\|\theta_0\|^2$ reparametrization factor. $\rho$ removes it, because both its numerator and its
denominator are measured at the same anchor. Cross-family comparisons therefore use $\rho$ alone.

\paragraph{Evaluation corpora.} Three corpora carry the measurements. We will release all three with the code.
\emph{Mixed}, the headline corpus, is the first 50{,}000 characters of each of four domain files: web
(FineWeb-Edu), chat (UltraChat-200k, \texttt{test\_sft} split), math (Hendrycks MATH, algebra), and code (CodeParrot clean,
validation). Under the Olmo tokenizer it holds 51{,}167 tokens (10{,}311 web, 9{,}868 chat, 18{,}325 math, 12{,}663 code).
The pooled mean weights every position equally, so math and code carry 31k of the 51k.
\emph{WikiText} is the first 40{,}000 characters of the WikiText-2 raw test split \citep{wikitext} with blank
lines removed, truncated to 1{,}536 tokens.
\emph{Plain-math} is a fixed 471-word file of ten short worked problems in plain prose, 700--900 tokens
depending on the tokenizer. Records are flattened into one text stream before slicing. FineWeb-Edu documents and CodeParrot files
(200--20{,}000 characters) keep their stream order. UltraChat conversations become \texttt{User:}/\texttt{Assistant:} lines, and
MATH problems become \texttt{Problem:}/\texttt{Solution:} pairs.
All directions at one anchor are measured on identical tokens. The per-position quadratic coefficient is
heavy-tailed. For the released deltas on the mixed corpus, the top 10\% of positions carry 43--77\% of the corpus
total and the top 1\% carries 11--44\%.

\paragraph{Finite-difference reads.} The window ladders of Tables~\ref{tab:windows_olmo} and~\ref{tab:windows_pythia} (Appendix~\ref{app:pythia}), the bf16
controlled runs of Table~\ref{tab:long} (Appendix~\ref{app:traj}), and the extrapolation tests (Appendix~\ref{app:astar}) come from an
earlier campaign that measured $c(\alpha) = \mathrm{KL}(\alpha)/(\alpha\relDW)^2$ by finite differences over
$\alpha \in \{0.003, 0.01, 0.03, 0.1, 0.3, 1.0\}$, with $\mathrm{KL}(\alpha) = \mathbb{E}_t\sum_y p_{t,y}(\theta_0 +
\alpha\Delta)\log[p_{t,y}(\theta_0 + \alpha\Delta)/p_{t,y}(\theta_0)]$, the perturbed model in the first argument
as in \S\ref{sec:method}. A scale $\alpha$ is certified for an item (a delta or one null
draw) on a corpus when it passes three tests. (i) Its mean per-token KL is at least $10^{-6}$ nats. (ii) The ratio of its KL to the
KL at the neighboring sampled scale lies within a factor $1.53$ above or $0.63$ below the quadratic value. (iii)
$c(\alpha)$ lies within the same band around $c$ at a reference scale whose KL exceeds $10^{-4}$. A read takes the delta and
every draw of a null kind at the largest scale certified for all of them, the certified common scale of Tables~\ref{tab:r4_released} and~\ref{tab:windows_olmo}. Chords at full norm measure destruction
rather than curvature: a random direction with a pretraining window's norm shows $c(\alpha)$ \emph{rising} toward $\alpha = 1$, and the
Pythia-160M step-32{,}000$\rightarrow$final chord peaks at KL $2.66$ nats at $\alpha = 0.3$ and falls back to $0.40$
at $\alpha = 1$, although both endpoints are real checkpoints. Every pretraining level therefore comes from
small-$\alpha$ probes. Olmo stage-1 checkpoints are stored in bf16. The rounding of a window's two endpoints accounts for at most $0.8\%$ of
$\|\Delta\|^2$ for the 814-step window and $0.13\%$ at 7{,}814 steps, so the narrow windows are weight motion, not
quantization noise. Pythia shards are float32.

\begingroup\raggedright
\paragraph{Checkpoints.} The Olmo-3-7B \citep{olmo3} base anchor is \texttt{allenai/Olmo-3-1025-7B} at
revision \texttt{main}. \texttt{pre\_late\_back} is the window from revision \texttt{stage1-step1272000}
to \texttt{main}. \texttt{pre\_mid\_back} is the window from \texttt{stage1-step1413814}, the last stage-1
checkpoint, to \texttt{main}. Both windows include the post-stage-1 training. The within-stage-1 windows of
\S\ref{sec:pretrain} are anchored at \texttt{stage1-step1413814} and run back to the released stage-1
checkpoints 814, 1{,}814, 2{,}814, 3{,}814, 4{,}814, 5{,}814, 7{,}814, 10{,}814, 30{,}814, 100{,}814,
200{,}814, 400{,}814, 700{,}814, and 1{,}000{,}814 steps earlier. The Instruct pipeline directions are
\texttt{allenai/Olmo-3-7B-Instruct-SFT} (from the base), \texttt{-Instruct-DPO} (from the SFT), and
\texttt{-Instruct} (from the DPO). The Think pipeline uses \texttt{-Think-SFT}, \texttt{-Think-DPO}, and
\texttt{-Think} in the same way. The five \texttt{allenai/Olmo-3-7B-RL-Zero-\{Math, Code, IF, General,
Mix\}} checkpoints start from the base. All are at revision \texttt{main}, and each delta is measured at its anchor. The T\"ulu-3 \citep{tulu3}
deltas are \texttt{allenai/Llama-3.1-Tulu-3-8B-SFT}$\rightarrow$\texttt{-DPO} and \texttt{-DPO}$\rightarrow$\texttt{-8B}
(the RLVR endpoint). The Zephyr \citep{zephyr} delta is
\texttt{HuggingFaceH4/mistral-7b-sft-beta}$\rightarrow$\texttt{zephyr-7b-beta}.
Pythia \citep{pythia} windows use the \texttt{stepN} revisions of \texttt{EleutherAI/pythia-1.4b} (anchors
\texttt{step143000} and \texttt{step100000}) and \texttt{pythia-160m}. The released Qwen2.5-0.5B-Instruct
delta is measured at \texttt{Qwen/Qwen2.5-0.5B}. The OLMo-2-0425-1B DPO delta is \texttt{allenai/OLMo-2-0425-1B-SFT}$\rightarrow$\texttt{-DPO}. The OLMo-2 RLVR
deltas run from \texttt{-DPO} to \texttt{-Instruct} at revisions \texttt{step\_200} to \texttt{step\_2600} and
\texttt{main}. The DARE deltas are Qwen2.5-0.5B and Qwen2.5-1.5B to their Instruct models, Qwen2.5-Math-1.5B to
Qwen2.5-Math-1.5B-Instruct and to DeepSeek-R1-Distill-Qwen-1.5B, and the OLMo-2-1B DPO and final RLVR deltas. The 21 deltas of the
extrapolation tests are listed in Appendix~\ref{app:astar}.\par\endgroup

\paragraph{Licenses.} Olmo-3, OLMo-2, Qwen2.5, Qwen2.5-Math, SmolLM2, Pythia, and Mistral-7B are released
under Apache-2.0 (Qwen2.5-3B under the Qwen Research License). DeepSeek-R1-Distill, DeepScaleR, Zephyr,
UltraFeedback-binarized, UltraChat, MATH, MATH-500, and GSM8K are under MIT. The OLMo-2 0425 1B preference mixture is under ODC-By-1.0.
Its subsets carry their own licenses, some non-commercial. T\"ulu-3 and Llama-3.1 are under the Llama 3.1 Community License.
ProRL (Nemotron-Research-Reasoning-Qwen-1.5B) is under CC-BY-NC-4.0.
WikiText is under CC-BY-SA-3.0, FineWeb-Edu under ODC-By-1.0, and CodeParrot under the licenses of its source
repositories. All are used for research evaluation within those terms.

\paragraph{Compute.} The exact-form measurements and training runs took about 115 wall-clock hours and 135 GPU-hours: one
A100 80GB for the 0.5B and 1B runs, their measurement, and the Pythia windows; one H100 for the leave-one-out and
constant-rate arms; one B200 for the 7B runs; one H200 for the 7--8B released deltas, the stage-1 windows, the
truncation ladder, and the extra Think-anchor draws; and two H100s for each 8.6-hour reproduction. A 7B forward-mode pass over the three corpora takes about 1.2 minutes
on an H200, including model loading. The earlier finite-difference campaigns behind Appendices~\ref{app:pythia},
\ref{app:traj}, and~\ref{app:astar} took about 120--140 GPU-hours on A100, H100, and H200 GPUs, including pilots
and re-runs that do not appear in the paper.

\FloatBarrier
\section{Pretraining windows: complete results}
\label{app:pythia}

\paragraph{Olmo-3-7B stage-1 windows, exact form.} Table~\ref{tab:r4_stage1} measures every window of Table~\ref{tab:windows_olmo} with
the exact form and per-delta nulls (6 isotropic, 3 profile, 3 sign-flip draws). On plain-math and WikiText the
exact and finite-difference reads agree to within 40\% from 30k steps on; at 814--7{,}814 steps the exact form reads
1.1--5 times higher on plain-math. On the mixed corpus the gap is larger and sits in a few positions. The top 1\% of
positions carry 57--81\% of the form at 814--7{,}814 steps and 21--72\% beyond, against 20\% for the nulls. The
pooled read therefore runs from 322 at 814 steps to 0.8--12 on the plateau, while the median tracks the finite-difference
read (5.2 to 0.7--1.0). The finite-difference reads below set these positions aside as a scale-dependent
excess at $\alpha \le 0.1$. The exact form shows that they carry second-order curvature, concentrated on math and code
text. Slopes of $\log\rho$ against $\log$ width over 814--7{,}814 steps are $-0.74$ (plain-math), $-0.52$
(WikiText), $-0.70$ (web), $-0.49$ (chat), $-1.73$ (math), and $-0.87$ (code). Geometric means over 30k--1M steps,
in the same order, are 1.10, 1.14, 0.82, 0.72, 3.5, and 2.1.

\begin{table}[H]
\centering
\scriptsize
\setlength{\tabcolsep}{3pt}
\resizebox{\linewidth}{!}{\input{tables/tab_r4_stage1}}
\caption{\textbf{Olmo-3-7B stage-1 windows, exact form against finite differences.} Backward windows anchored at
\texttt{stage1-step1413814}; isotropic leverage by the exact form and by the earlier finite-difference read (FD)
on the two small corpora, per domain and pooled on the mixed corpus (mean and median), and the exact sign-flip
read and magnitude alignment on the mixed corpus.}
\label{tab:r4_stage1}
\end{table}

\paragraph{Olmo-3-7B, end-anchored, finite differences.} Table~\ref{tab:windows_olmo} lists every backward window at the last
stage-1 checkpoint. Windows of $7{,}814$ steps or fewer are read at $\alpha = 0.3$ for both delta and nulls.
At $\alpha \le 0.1$ a few positions (top-1\% share 0.35--0.55) add a roughly constant 0.2--2.4 nats to the summed
KL of these deltas. This excess is absent from every null draw, appears at one scale and not its neighbors, and is
$5$--$200\times$ the quadratic term there. At $\alpha = 0.3$ the quadratic term (4--5 nats on plain-math) swamps it,
and the Pythia deltas show no such excess. At $\alpha = 1$ these
deltas and their nulls are already super-quadratic ($c(1.0)/c(0.3)$ rises from $1.07$ at 814 steps to $2.1$ at 7{,}814). The certified
chain accepts $\alpha = 1$ on WikiText and the mixed corpus, where the $\alpha = 1$ read exceeds the $\alpha = 0.3$ read
by 10--26\%; we do not use it. Wider windows are read at the certified common scale. Fitted slopes of
$\log\rho$ against $\log$ width over 814--7{,}814 steps are $-0.53$, $-0.48$, $-0.47$ (plain-math, WikiText, mixed;
$R^2 = 0.98$--$0.99$), and $-1.2$ to $-1.3$ against $\log\relDW$. Over 814--30{,}814 steps the slopes are $-0.44$,
$-0.42$, $-0.40$. Over 30k--1M steps they are $-0.09$ to $+0.01$ with $R^2 \le 0.46$. Geometric means over the 30k--1M
plateau are $1.09$--$1.18$ (isotropic), $1.31$--$1.42$ (profile), and $1.23$--$1.59$ (sign-flip). The
median-based isotropic read crosses 1 at 13k (WikiText) and 18k (mixed) steps and then stays at $0.85$--$1.5$.
Displacement grows as width$^{0.35\text{--}0.42}$ below 10k steps and width$^{0.32\text{--}0.36}$ to 100k.
Beyond 400k it grows linearly (width$^{0.9\text{--}1.2}$); there the window reaches back into the early
high-learning-rate part of the run. The narrow deltas are 64\% MLP, 26\% attention, and 10\% embeddings by
squared norm; the embedding share rises to 43\% at 1M steps. The domain order on the mixed corpus, math $>$ code $>$ web $>$ chat, holds at
every width to 200k steps (Table~\ref{tab:r4_stage1}).

\begin{table}[H]
\centering
\input{tables/tab_windows_olmo}
\caption{\textbf{Olmo-3-7B backward windows at the last stage-1 checkpoint}, every width from 814 to
1{,}000{,}814 steps, leverage by finite differences against the three per-delta nulls on the three corpora. ``read'': both sides at
$\alpha = 0.3$ (narrow windows) or at the certified common scale.}
\label{tab:windows_olmo}
\end{table}

\paragraph{Pythia-1.4B, exact form.} Table~\ref{tab:r5_pythia} measures the end-anchored windows with the exact
form (6 isotropic, 3 profile, 3 sign-flip draws). The reads sit 10--46\% below the finite-difference values, and the gap widens with
width. The top 1\% of positions carry 11\% of the form (16\% at 43k steps), and the magnitude alignment is
0.26--0.37, so Adam's tilt toward flat coordinates (Appendix~\ref{app:nqm}) is visible in this lineage at every
width.

\begin{table}[H]
\centering
\scriptsize
\setlength{\tabcolsep}{3pt}
\resizebox{\linewidth}{!}{\input{tables/tab_r5_pythia}}
\caption{\textbf{Pythia-1.4B windows at step 143{,}000, exact form against finite differences} (FD: the end-anchored
certified-scale read of Table~\ref{tab:windows_pythia}).}
\label{tab:r5_pythia}
\end{table}

\paragraph{Pythia-1.4B, both conventions.} Table~\ref{tab:windows_pythia} gives the same widths measured
backward from step 143{,}000 (the window's end) and forward from step 100{,}000 (the window's start). The end-anchored read is $1.5$--$2.0\times$ the start-anchored one at every width from
1k to 16k (delta curvature $2.4$--$3.5\times$, null curvature $1.5$--$1.75\times$); the backward window of a
given width has $0.47$--$0.64\times$ the displacement of the forward window, because the learning rate decays
over this stretch. At 43k steps the two conventions share one delta (step 100{,}000 to 143{,}000) and differ
only in the anchor. Slopes of $\log\rho$ against $\log$ width are $-0.53$, $-0.58$, $-0.47$ end-anchored
($R^2 = 0.84$--$0.95$) and $-0.71$, $-0.72$, $-0.72$ start-anchored ($R^2 = 1.00$). Against $\log\relDW$ they
are $-0.86$ to $-0.94$ end-anchored and $-1.84$ to $-1.87$ start-anchored. The end-anchored ladder crosses
$\rho = 1$ at 10.7k, 12.8k, and 13.2k steps ($\relDW = 0.037$--$0.042$) against 5.1k--6.3k steps
($\relDW = 0.049$--$0.053$) start-anchored. The sign-flip null gives the highest read at narrow
widths in both lineages (Olmo 814 steps: $8$--$11$; Pythia 1k steps: $13$--$16$), about twice the isotropic and profile
reads, and converges to them as the window widens.

\begin{table}[H]
\centering
\input{tables/tab_windows_pythia}
\caption{\textbf{Pythia-1.4B windows measured at their end (step 143{,}000) and at their start (step
100{,}000)}, isotropic leverage per corpus with the end/start ratio, and the profile and sign-flip reads of
the end-anchored windows on the mixed corpus. The 43k row is one delta at two anchors.}
\label{tab:windows_pythia}
\end{table}

\subsection{A noisy quadratic model of window leverage}
\label{app:nqm}
Take a quadratic loss with Hessian $H = F = \mathrm{diag}(\lambda_1, \ldots, \lambda_d)$ and SGD with step $\eta$,
batch $B$, and gradient noise covariance $C = F/B$. In the continuous-time limit the stationary covariance
solves $H\Sigma + \Sigma H = \eta C$ \citep[Eq.~13]{mandt}, so $\Sigma = (\eta/2B)\,I$. The displacement over a window
of width $w$ in the stationary regime has covariance $\mathbb{E}[\Delta\Delta^\top] = 2\Sigma - \Sigma e^{-\eta wF} -
e^{-\eta wF}\Sigma = (\eta/B)(I - e^{-\eta wF})$. With $s_k = 1 - e^{-\eta w\lambda_k}$, the ratio of expectations is
\[
\rho(w) = \frac{\mathbb{E}[\Delta^\top F\Delta] / \mathbb{E}\|\Delta\|^2}{\tfrac1d\operatorname{tr}F}
= \frac{\sum_k \lambda_k s_k / \sum_k s_k}{\tfrac1d\sum_k\lambda_k}, \qquad
\mathbb{E}\|\Delta\|^2 = \frac{\eta}{B}\sum_k s_k .
\]
The expectation of the Rayleigh quotient itself agrees with this ratio once many modes are active. At $w \to 0$ the
top modes dominate and the two differ.
As $w \to \infty$, $s_k \to 1$ and $\rho \to 1$; as $w \to 0$, $s_k \approx \eta w\lambda_k$ and $\rho \to
d\sum_k\lambda_k^2/(\sum_k\lambda_k)^2$. For a diagonal preconditioner $P$ the stationary equation becomes
$P^{-1}F\Sigma + \Sigma FP^{-1} = \eta P^{-1}CP^{-1}$. With $C = F/B$, $\Sigma = (\eta/2B)\,P^{-1}$ solves it exactly
for any symmetric positive-definite $F$, diagonal or not. With $P_{ii} = \sqrt{F_{ii}/B}$
(Adam's second-moment estimate near the optimum) this gives $\Sigma_{ii} = \eta/(2\sqrt{BF_{ii}})$. Because
$\Sigma$ is diagonal, $\rho_\infty = \overline{F^{1/2}}\,/\,(\overline{F}\;\overline{F^{-1/2}}) \le 1$ by Chebyshev's
sum inequality, where $\overline{f(F)} = \tfrac1d\sum_i f(F_{ii})$. The finite-width covariance is
$\mathbb{E}[\Delta\Delta^\top] = (\eta/B)\,P^{-1/2}(I - e^{-\eta w\tilde F})P^{-1/2}$ with $\tilde F = P^{-1/2}FP^{-1/2}$.
It contains both regimes. At narrow widths it is $(\eta^2 w/B)\,P^{-1}FP^{-1}$, whose off-diagonal part gives
$\rho_{\text{sign}} > 1$ whenever $F$ is not diagonal. At wide widths it is $2\Sigma$. Even for a diagonal $F$ the
coordinates do not equilibrate at one rate. Coordinate $i$ has $\mathbb{E}[\Delta_i^2] = (\eta/\sqrt{BF_{ii}})
(1 - e^{-\eta w\sqrt{BF_{ii}}})$. Every coordinate starts with the same variance $\eta^2 w$ ($\rho = 1$), and the
steep ones saturate first; this saturation carries $\rho$ down to $\rho_\infty$. The measured narrow windows read
$\rho_{\text{sign}} = 616$ (Olmo, 814 steps) and 14 (Pythia, 1{,}000 steps) by the exact form and 8--16 by finite
differences, with a magnitude alignment below 1 (Tables~\ref{tab:r4_stage1} and~\ref{tab:r5_pythia}). The
narrow-window excess therefore sits in cross-coordinate modes that a diagonal preconditioner does not whiten. The
transition that sets the width dependence is the one in the eigenbasis of $F$, not the coordinate-wise saturation. Table~\ref{tab:nqm} evaluates $\rho(w)$ and $\|\Delta\|$ for $\lambda_k = k^{-a}$ ($\lambda_{\max} = 1$),
$d = 2\times10^6$, fitting exponents over $\eta w \in [30, 10^4]$ (\texttt{experiments/theory/nqm\_leverage.py}).
For $1/2 < a < 1$ and $1 \ll \eta w \ll d^a$ the exponents tend to $1/a - 2$ for $\rho$ ($-0.33$, $-0.50$, $-0.67$,
and $-0.89$ for the four rows) and to $1/2$ for $\|\Delta\|$, because the unequilibrated tail of the spectrum
dominates $\sum_k s_k$. The fitted values sit between those limits and zero, because at $d = 2\times10^6$ the cutoff
$d^a$ ($6\times10^3$ to $5\times10^5$) lies inside or next to the fit window. At $d = 2\times10^7$ the $\rho$ exponents move to
$-0.33$ to $-0.73$ and the $\|\Delta\|$ exponents to $0.45$--$0.47$. Here $d$ stands for the number of modes that equilibrate over the window
ladder, not for the parameter count. At $\eta w = 1$ the top modes have already partly equilibrated, so the first column sits 14--24\% below the
$w \to 0$ limit; at $a = 0.90$ the flattest modes ($\lambda_d \approx 2\times10^{-6}$) have not equilibrated by
$\eta w = 10^5$, which is why that row still reads 2.0 there.

\begin{table}[H]
\centering
{\small\setlength{\tabcolsep}{4pt}\begin{tabular}{rrrrrr}
\toprule
$a$ & $\rho$ at $\eta w = 1$ & inv.\ participation ratio & $\rho$ at $\eta w = 10^5$ & $d\log\rho/d\log w$ & $d\log\|\Delta\|/d\log w$ \\
\midrule
0.60 & 13.4 & 15.6 & 1.00 & $-0.29$ & 0.38 \\
0.67 & 41.7 & 50.7 & 1.00 & $-0.41$ & 0.42 \\
0.75 & 191 & 242 & 1.04 & $-0.54$ & 0.44 \\
0.90 & 2{,}580 & 3{,}410 & 2.03 & $-0.70$ & 0.45 \\
\bottomrule
\end{tabular}}
\caption{\textbf{The noisy quadratic model} for power-law spectra $\lambda_k = k^{-a}$: leverage at one unit of
$\eta w$, its $w \to 0$ limit (the inverse participation ratio), its value at $\eta w = 10^5$, and the fitted
exponents of $\rho$ and $\|\Delta\|$ in width.}
\label{tab:nqm}
\end{table}

\FloatBarrier
\section{Training length}
\label{app:traj}
With fp32 master weights, the 10{,}000-step SFT and DPO runs of Table~\ref{tab:r4_controlled} overlap the 600-step
runs at the same rate ($5\times10^{-6}$), seed, and data order over $\relDW = 4.3$--$6.0\times10^{-3}$ (SFT) and $1.9$--$2.9\times10^{-3}$ (DPO).
Interpolated on the mixed corpus, SFT reads 8.7 (long) against 8.8 (short) at $5\times10^{-3}$ and DPO 115 against
114 at $2\times10^{-3}$; the long runs then continue to $\rho = 3.5$ at $\relDW = 1.7\times10^{-2}$ (slope $-0.73$ against $-1.38$ over the first 600 steps)
and 31 at $1.1\times10^{-2}$ ($-0.85$ against $-0.70$), with magnitude alignment 0.42 and 0.73 at their ends. Displacement grows
through the whole run (SFT $6.0\times10^{-3}$ at step 600 to $1.7\times10^{-2}$ at 10{,}000), unlike the bf16 SFT runs
below, whose displacement saturated.
An earlier finite-difference campaign had trained the same base with bf16 parameters and no fp32 master copy, so
its levels, slopes, and displacements carry the bf16 bias of \S\ref{sec:repro}. The body does not use them. Within
that setting, its training-length comparison gave the same answer. SFT at $5\times10^{-7}$ and $5\times10^{-6}$ and DPO at
$5\times10^{-6}$ and $5\times10^{-5}$ ran for 10{,}000 steps (1.3 epochs of the 61{,}135 UltraFeedback-binarized
training pairs), and GRPO \citep{grpo} at $1\times10^{-6}$ and $3\times10^{-6}$ for 3{,}000 and 1{,}000 steps on
GSM8K, against the campaign's shorter runs (600 steps, two of them 2{,}400; 150 for GRPO) at rates from $5\times10^{-7}$ to $5\times10^{-5}$. Training used TRL \citep{trl} with AdamW
$(0.9, 0.999, 10^{-8})$, weight decay 0, gradient clipping at 1.0, a constant learning rate, and batch 8. DPO used
the sigmoid loss with $\beta = 0.1$. GRPO used four generations per prompt and a reward for the correct final number.
We read curvature by finite differences at a certified scale on plain-math and WikiText, and divided both by one
fixed isotropic null, $c = 95$, measured on plain-math (four draws at $\relDW \in \{10^{-3}, 10^{-2}, 10^{-1}\}$ gave 95, 92, and 86).
At matched displacement those long runs read at most 13\% above the short ones, and usually below them
(Table~\ref{tab:long}).

\begin{table}[H]
\centering
\input{tables/tab_long}
\caption{\textbf{Training length at matched displacement in the earlier bf16 runs}: geometric-mean $\rho = c/95$ per
half-decade bin for the short runs (600 steps, two of them 2{,}400; 150 for GRPO) and the long runs (10{,}000 steps; 3{,}000 and 1{,}000
for GRPO), $n$ checkpoints in parentheses.}
\label{tab:long}
\end{table}

\FloatBarrier
\section{Prospective tests of extrapolation tolerance}
\label{app:astar}

\paragraph{The $\astar$ test (first test).} The quadratic radius $\astar$ comes from two passes over the
finite-difference $c(\alpha) = \mathrm{KL}(\alpha)/(\alpha\relDW)^2$ at the sampled scales (0.01 to 1.0, skipping
scales whose KL is zero). A provisional $c_0$ is the geometric mean over $\alpha \le 0.1$. Then $\astar$ is the largest
sampled $\alpha$ with $c(\alpha)/c_0 \in [0.7, 1.43]$, or the smallest sampled $\alpha$ if none qualifies. The
final $c_0$ is the geometric mean over $\alpha \le \min(\astar, 0.1)$. For the 17 released deltas of Table~\ref{tab:astar} we
measured $\astar$,
applied a rule written in advance (aggregate $\astar$ as the minimum over corpora; \textsc{tolerant} if
$1.0$, \textsc{fragile} if $\le 0.3$; a \textsc{tolerant} delta should keep at least half its endpoint gain
at $1.5\times$ and a \textsc{fragile} delta less than half), and committed the predictions to a timestamped repository, to be released with the code, before any benchmark ran.
Scoring used GSM8K \citep{gsm8k}: the first 150 test problems, 4-shot with the first four training problems in a
Question/Answer template, greedy decoding to 256 tokens. An answer is correct when the number after ``The answer is'',
or else the last number, matches the gold answer. A delta is unscorable when $a_1 \le a_0$, with $a_\alpha$ the accuracy at scale $\alpha$. The rule scored
5 of 8 scorable deltas (binomial $p = 0.36$, CI $[0.245, 0.915]$). The post hoc rule $\relDW < 10^{-2}$ scored 7 of 8 on the same outcomes. Under that rule \texttt{qwen15\_instruct}, $3\%$ above the
threshold, is \textsc{fragile} and counts as a hit; the one miss is \texttt{olmo\_rlzero\_math}. Intervals here and below are
two-sided 95\% Clopper--Pearson intervals; the $p$-values are one-sided.

\begin{table}[H]
\centering
{\footnotesize\setlength{\tabcolsep}{3.5pt}
\resizebox{\linewidth}{!}{%
\begin{tabular}{llrrrrlrrrrl}
\toprule
family & delta & $\relDW$ & $\astar$ wiki & $\astar$ math & $\astar$ & prediction & $a_0$ & $a_1$ & $a_{1.5}$ & kept & outcome \\
\midrule
DeepScaleR & \texttt{deepscaler\_rlvr} & 8.10e-04 & 1.00 & 1.00 & 1.00 & \textsc{tolerant} & 0.740 & 0.693 & 0.553 & -- & unscorable \\
\addlinespace[2pt]
Llama/T\"ulu-3 & \texttt{tulu\_rlvr} & 7.07e-04 & 1.00 & 1.00 & 1.00 & \textsc{tolerant} & 0.827 & 0.820 & 0.820 & -- & unscorable \\
Llama/T\"ulu-3 & \texttt{tulu\_dpo} & 8.40e-04 & 0.30 & 0.30 & 0.30 & \textsc{fragile} & 0.740 & 0.827 & 0.860 & 1.38 & miss \\
\addlinespace[2pt]
Mistral/Zephyr & \texttt{zephyr\_dpo} & 1.40e-03 & 1.00 & 1.00 & 1.00 & \textsc{tolerant} & 0.373 & 0.367 & 0.233 & -- & unscorable \\
\addlinespace[2pt]
Olmo-3-7B & \texttt{olmo\_dpo\_think} & 3.81e-04 & 1.00 & 1.00 & 1.00 & \textsc{tolerant} & 0.833 & 0.873 & 0.873 & 1.00 & hit \\
Olmo-3-7B & \texttt{olmo\_rlzero\_gen} & 6.39e-04 & 1.00 & 1.00 & 1.00 & \textsc{tolerant} & 0.687 & 0.713 & 0.720 & 1.25 & hit \\
Olmo-3-7B & \texttt{olmo\_rlvr\_think} & 8.06e-04 & 1.00 & 1.00 & 1.00 & \textsc{tolerant} & 0.867 & 0.867 & 0.853 & -- & unscorable \\
Olmo-3-7B & \texttt{olmo\_dpo\_instruct} & 8.84e-04 & 1.00 & 1.00 & 1.00 & \textsc{tolerant} & 0.760 & 0.707 & 0.633 & -- & unscorable \\
Olmo-3-7B & \texttt{olmo\_rlzero\_mix} & 8.98e-04 & 1.00 & 1.00 & 1.00 & \textsc{tolerant} & 0.687 & 0.727 & 0.760 & 1.83 & hit \\
Olmo-3-7B & \texttt{olmo\_rlvr\_instruct} & 1.08e-03 & 1.00 & 1.00 & 1.00 & \textsc{tolerant} & 0.707 & 0.700 & 0.680 & -- & unscorable \\
Olmo-3-7B & \texttt{olmo\_rlzero\_math} & 1.27e-03 & 1.00 & 1.00 & 1.00 & \textsc{tolerant} & 0.687 & 0.740 & 0.700 & 0.25 & miss \\
Olmo-3-7B & \texttt{olmo\_sft\_think} & 1.50e-01 & 0.03 & 0.10 & 0.03 & \textsc{fragile} & 0.687 & 0.827 & 0.307 & -2.71 & hit \\
Olmo-3-7B & \texttt{olmo\_sft\_instruct} & 1.62e-01 & 0.03 & 0.10 & 0.03 & \textsc{fragile} & 0.687 & 0.800 & 0.400 & -2.53 & hit \\
\addlinespace[2pt]
ProRL & \texttt{prorl\_rlvr} & 3.20e-03 & 0.30 & 0.30 & 0.30 & \textsc{fragile} & 0.740 & 0.607 & 0.333 & -- & unscorable \\
\addlinespace[2pt]
Qwen/R1-Distill & \texttt{r1distill\_sft} & 1.19e-01 & 1.00 & 0.10 & 0.10 & \textsc{fragile} & 0.720 & 0.080 & 0.000 & -- & unscorable \\
\addlinespace[2pt]
Qwen2.5-1.5B & \texttt{qwen15\_instruct} & 1.03e-02 & 1.00 & 1.00 & 1.00 & \textsc{tolerant} & 0.620 & 0.633 & 0.613 & -0.50 & miss \\
Qwen2.5-Math & \texttt{qwenmath\_instruct} & 1.57e-01 & 0.30 & 1.00 & 0.30 & \textsc{fragile} & 0.720 & 0.707 & 0.313 & -- & unscorable \\
\bottomrule
\end{tabular}}}
\caption{\textbf{The first prospective test} ($\astar$ rule, GSM8K, 150 problems, 17 released deltas): quadratic
radius per corpus, prediction, accuracy at $0\times$, $1\times$, and $1.5\times$, the fraction of the endpoint gain kept,
and the outcome.}
\label{tab:astar}
\end{table}

\paragraph{The displacement rule (second test).} The rule: a delta with $\relDW < 10^{-2}$ keeps
at least half its endpoint gain at $1.5\times$ (\textsc{tolerant}); a delta with $\relDW \ge 10^{-2}$ loses
more than half (\textsc{fragile}). Scoring was fixed in the registration. A delta is scorable iff $a_1 - a_0 \ge 0.10$ on
200 problems; kept $= (a_{1.5} - a_0)/(a_1 - a_0)$; \textsc{tolerant} is correct iff kept $\ge 0.5$ and
\textsc{fragile} iff kept $< 0.5$. The headline is correct/scorable against a binomial null of $0.5$
(one-sided $p$, 95\% Clopper--Pearson interval); the same count at $2.0\times$ is secondary. Blends
are $\theta_{\text{anchor}} + \alpha\Delta$ computed in float32 and cast to bf16, $\alpha \in \{0, 1, 1.5,
2\}$. Generation uses the anchor's tokenizer, architecture, and end-of-sequence ids and no chat template.
\emph{GSM8K} (test, first 200 problems): 4-shot with the first four training problems in a Question/Answer
template, greedy, 512 new tokens, answer $=$ the number after ``The answer is'' or else the last number,
stopping at ``Question:''. Twelve fresh deltas were registered in two parts (eight and four) at 2026-10-06
11:24:56 and 12:18:51 UTC (repository commits \texttt{bfeebd7f} and \texttt{fe311b5e}), each part before any
benchmark on its deltas. \emph{MATH-500} (first 200 problems in dataset order): 4-shot Minerva-style prompt
with four fixed level-2--3 training problems whose solutions end in \verb|\boxed{}|, greedy, 768 new tokens,
stopping at ``Problem:''; answer $=$ the last \verb|\boxed{}| (no box counts as wrong); comparator math-verify 0.9.0,
with a normalizer-based comparator stored beside it (the normalizer changes no scorable verdict). Twenty-one deltas (the twelve
fresh ones plus nine the first test could not score) were registered at 2026-10-08 00:09:13 UTC (commit
\texttt{5f6f2cb7}), with no code change after registration. Both registrations fix every prediction from
$\relDW$ values measured before the experiment.

\paragraph{Results.} GSM8K: 3 of 3 scorable correct at $1.5\times$ ($p = 0.125$, CI $[0.29, 1.00]$), and the same
three at $2.0\times$. Nine of 12 are unscorable because the anchor already answers 4-shot GSM8K within six points of
the endpoint. MATH-500: 1 of 5 at $1.5\times$ ($p = 0.969$, CI $[0.005, 0.716]$) and 3 of 5 at $2.0\times$. With the normalizer comparator the count is 1 of 4, because
\texttt{tulu31\_rlvr}'s gain of exactly 0.100 falls below the scorable line; no scorable verdict changes. Sixteen of 21 are unscorable, ten of them because the endpoint scores below its
anchor under the raw prompt; the R1-distilled endpoint scores 41 points below and reaches a boxed answer in 10\% of
completions within 768 tokens. Pooled over distinct deltas, taking the MATH verdict where a delta is scorable on both: 2 of 6
($p = 0.89$). The post hoc $\astar$ rule (\textsc{tolerant} iff $\astar = 1.0$) scores 2 of 3 on GSM8K and 3
of 5 on MATH. It reads $\astar = 1.0$, hence \textsc{tolerant}, for 15 of 21 deltas. Descriptive mean changes from the anchor over all deltas keep the rule's direction
clearly only at $2\times$ (\textsc{fragile}-labeled $-12.7$ points against $-4.2$ for \textsc{tolerant}-labeled on
MATH; $-18.9$ against $-4.3$ on GSM8K). No threshold on $\relDW$ separates the MATH-scorable deltas: the
four that kept at least half their gain sit at $\relDW = 8.8\times10^{-4}$, $1.3\times10^{-2}$, $4.2\times10^{-2}$, and
$6.3\times10^{-2}$ and the one that did not at $2.0\times10^{-3}$. Table~\ref{tab:extrap} lists every delta.

\paragraph{Format drift.} At $1.5\times$ the three MATH-scorable instruct deltas rarely stop at the next ``Problem:'' (stop rate
$0.005$--$0.18$ against $0.22$--$0.69$ at $1.0\times$), and their GSM8K completions drop the ``The answer
is'' template (for \texttt{smollm2\_instruct}, present in 182 of 200 completions at $1.0\times$ and 20 of 200
at $1.5\times$) while still reasoning correctly, so the last-number fallback fails where MATH's last-box rule
survives trailing drift; the same delta lost its whole GSM8K gain at $1.5\times$ and kept 71\% of its MATH
gain. Extension can also help under a raw prompt: on MATH, \texttt{qwen7b\_instruct} reaches 0.68 at $1.5\times$ and
$2\times$ from 0.45 at the endpoint (anchor 0.49), and \texttt{qwen3b\_instruct} at $1.5\times$ beats its endpoint. For
deltas that differ from the anchor by about one bf16 ulp per element, $1.5\times$ rounds to 1 or 2 ulps, so
the smallest deltas' $1.5\times$ blends are coarse.

\begin{table}[H]
\centering
\input{tables/tab_extrap}
\caption{\textbf{The preregistered displacement rule on 21 released deltas.} GSM8K columns from the displacement-rule
test (new, 200 problems) or the $\astar$ test (earlier, 150 problems); MATH-500 columns from the second test. ``stop'' is the fraction of MATH-500 completions that stop at the next few-shot block at $1.0\times$ and $1.5\times$;
``$\astar$'' is the post hoc quadratic radius. The GSM8K verdict column applies to the twelve deltas of the second test;
earlier rows show the first test's accuracies and are marked unscorable in that column. ``kept'' is computed from exact
counts before the accuracies are rounded. Deltas are
anchor$\rightarrow$endpoint pairs: OLMo-2-7B SFT/DPO/Instruct, Olmo-3-7B pipelines and RL-Zero, T\"ulu-3
and 3.1, Zephyr, Mistral-7B-sft-beta, Qwen2.5-0.5B/3B/7B-Instruct, SmolLM2-1.7B-Instruct, Llama-3.1-T\"ulu-3-SFT,
R1-Distill-Qwen-1.5B \citep{r1}, DeepScaleR \citep{deepscaler}, ProRL \citep{prorl}, and Qwen2.5-Math-1.5B-Instruct \citep{qwenmath}.}
\label{tab:extrap}
\end{table}

\end{document}

%% file: tables/tab_r4_spectrum.tex
\begin{tabular}{lrlllll}
\toprule
stage & $n$ & $\relDW$ & $\rho_{\text{iso}}$ & $\rho_{\text{profile}}$ & $\rho_{\text{sign}}$ & $c_{\text{sign}}/c_{\text{profile}}$ \\
\midrule
pretraining windows & 2 & 3.7e-1--3.8e-1 & 0.86 (0.78--0.95) & 0.87 (0.77--0.99) & 0.87 (0.82--0.92) & 1.0 (0.93--1.1) \\
SFT & 2 & 1.5e-1--1.6e-1 & 5.5 (5.4--5.7) & 4.1 (4.0--4.3) & 3.2 (3.0--3.5) & 1.3 (1.2--1.3) \\
RL-Zero & 5 & 6.4e-4--1.5e-3 & 708 (388--1{,}508) & 837 (442--1{,}630) & 616 (354--1{,}226) & 1.4 (1.1--1.6) \\
RLVR after DPO & 3 & 7.1e-4--1.1e-3 & 553 (410--665) & 607 (519--725) & 408 (264--533) & 1.5 (1.2--2.0) \\
DPO & 4 & 3.8e-4--1.4e-3 & 3{,}632 (378--22{,}570) & 4{,}088 (392--21{,}557) & 1{,}702 (489--9{,}844) & 2.4 (0.8--5.3) \\
\midrule
OLMo-2-1B DPO & 1 & 8.5e-4 & 1{,}452 & 1{,}287 & 1{,}012 & 1.3 \\
OLMo-2-1B RLVR & 1 & 5.3e-4 & 563 & 469 & 474 & 0.99 \\
\bottomrule
\end{tabular}

%% file: tables/tab_r4_released_full.tex
\begin{tabular}{lrrrrrrrrr}
\toprule
direction & $\relDW$ & unchanged & $\rho_{\text{iso}}$ math & wiki & mixed & FD mixed & $\rho_{\text{profile}}$ & $\rho_{\text{sign}}$ & $c_{\text{sign}}/c_{\text{profile}}$ \\
\midrule
\texttt{pre\_late\_back} & 3.83e-1 & 0.00 & 1.0 & 1.1 & 0.95 & 0.46 & 0.99 & 0.92 & 1.08 \\
\texttt{pre\_mid\_back} & 3.71e-1 & 0.00 & 0.84 & 0.81 & 0.78 & 0.41 & 0.77 & 0.82 & 0.93 \\
\texttt{sft\_instruct} & 1.62e-1 & 0.01 & 5.1 & 10 & 5.7 & 2.0 & 4.3 & 3.5 & 1.22 \\
\texttt{sft\_think} & 1.50e-1 & 0.01 & 6.7 & 13 & 5.4 & 2.8 & 4.0 & 3.0 & 1.34 \\
\texttt{rlzero\_math} & 1.27e-3 & 0.58 & 4{,}356 & 235 & 710 & 707 & 865 & 544 & 1.59 \\
\texttt{rlzero\_code} & 1.51e-3 & 0.53 & 937 & 186 & 388 & 384 & 442 & 354 & 1.25 \\
\texttt{rlzero\_if} & 1.38e-3 & 0.55 & 703 & 345 & 419 & 519 & 529 & 463 & 1.14 \\
\texttt{rlzero\_general} & 6.39e-4 & 0.72 & 2{,}844 & 802 & 1{,}508 & 2{,}026 & 1{,}630 & 1{,}226 & 1.33 \\
\texttt{rlzero\_mix} & 8.98e-4 & 0.67 & 1{,}978 & 530 & 1{,}018 & 904 & 1{,}245 & 809 & 1.54 \\
\texttt{dpo\_instruct} & 8.84e-4 & 0.67 & 18{,}772 & 4{,}091 & 6{,}165 & 6{,}895 & 5{,}776 & 1{,}625 & 3.55 \\
\texttt{rlvr\_instruct} & 1.08e-3 & 0.61 & 1{,}504 & 446 & 665 & 660 & 725 & 533 & 1.36 \\
\texttt{dpo\_think} & 3.81e-4 & 0.79 & 7{,}607 & 4{,}740 & 3{,}304 & -- & 5{,}727 & 1{,}072 & 5.34 \\
\texttt{rlvr\_think} & 8.06e-4 & 0.68 & 661 & 578 & 410 & 345 & 519 & 264 & 1.96 \\
\texttt{dpo\_zephyr} & 1.40e-3 & 0.26 & 745 & 235 & 378 & 423 & 392 & 489 & 0.80 \\
\texttt{dpo\_tulu} & 8.40e-4 & 0.76 & 24{,}464 & 6{,}192 & 22{,}570 & 17{,}969 & 21{,}557 & 9{,}844 & 2.19 \\
\texttt{rlvr\_tulu} & 7.07e-4 & 0.79 & 1{,}476 & 112 & 621 & 590 & 593 & 480 & 1.24 \\
\texttt{olmo2\_1b\_dpo} & 8.52e-4 & 0.48 & 1{,}247 & 733 & 1{,}452 & -- & 1{,}287 & 1{,}012 & 1.27 \\
\texttt{olmo2\_1b\_rlvr\_final} & 5.34e-4 & 0.60 & 1{,}354 & 77 & 563 & -- & 469 & 474 & 0.99 \\
\bottomrule
\end{tabular}

%% file: tables/tab_r4_controlled.tex
\begin{tabular}{llllrlrrrl}
\toprule
recipe & arm & lr & $\relDW$ & unchanged & $\rho_{\text{iso}}$ (mixed) & slope mixed & math & wiki & $c_{\text{sign}}/c_{\text{profile}}$ \\
\midrule
fp32, b8, const. & SFT & 5e-7 & 2.4e-4--1.3e-3 & 0.00 & 267$\to$35 & -1.29 & -0.94 & -1.23 & 0.37--1.38 \\
fp32, b8, const. & SFT & 5e-6 & 1.4e-3--6.0e-3 & 0.00 & 59$\to$8.0 & -1.38 & -1.29 & -1.20 & 0.37--0.64 \\
fp32, b8, const. & SFT & 5e-5 & 1.2e-2--3.7e-2 & 0.00 & 21$\to$6.7 & -0.99 & -1.23 & -0.93 & 0.60--0.80 \\
fp32, b8, const. & DPO & 5e-7 & 9.4e-5--3.2e-4 & 0.00 & 728$\to$509 & -0.36 & +0.61 & +0.17 & 1.30--1.37 \\
fp32, b8, const. & DPO & 5e-6 & 9.1e-4--2.9e-3 & 0.00 & 198$\to$92 & -0.70 & -1.34 & -0.46 & 0.97--1.11 \\
fp32, b8, const. & DPO & 5e-5 & 9.1e-3--3.0e-2 & 0.00 & 65$\to$25 & -0.83 & -0.66 & -0.51 & 0.73--0.97 \\
bf16 params, b8, const. & SFT & 5e-7 & 3.2e-5--1.9e-4 & 0.99 & 23$\to$35 & +0.27 & +0.19 & +0.22 & 9.82--14.31 \\
bf16 params, b8, const. & SFT & 5e-6 & 5.5e-4--3.9e-3 & 0.90 & 47$\to$2.5 & -1.51 & -1.44 & -1.51 & 0.38--1.87 \\
bf16 params, b8, const. & SFT & 5e-5 & 8.9e-3--2.1e-2 & 0.36 & 11$\to$5.4 & -0.89 & -1.27 & -0.93 & 0.70--0.89 \\
bf16 params, b8, const. & DPO & 5e-7 & 6.6e-6--2.3e-5 & 0.99 & 25$\to$51 & +0.56 & +0.38 & +0.47 & 5.55--7.10 \\
bf16 params, b8, const. & DPO & 5e-6 & 2.0e-4--5.9e-4 & 0.91 & 90$\to$68 & -0.21 & +0.10 & +0.20 & 3.87--5.30 \\
bf16 params, b8, const. & DPO & 5e-5 & 6.0e-3--1.7e-2 & 0.37 & 58$\to$18 & -1.06 & -0.80 & -0.86 & 0.92--1.19 \\
fp32, b8, linear & SFT & 5e-6 & 7.6e-4--4.6e-3 & 0.00 & 117$\to$7.2 & -1.55 & -1.39 & -1.44 & 0.29--0.66 \\
fp32, b8, linear & DPO & 5e-6 & 3.5e-4--1.7e-3 & 0.00 & 354$\to$114 & -0.72 & -0.42 & -0.58 & 0.99--1.08 \\
fp32, b8, linear & DPO & 5e-5 & 3.4e-3--1.7e-2 & 0.00 & 138$\to$31 & -0.91 & -1.06 & -0.80 & 0.73--0.90 \\
fp32, b128, const. & SFT & 5e-6 & 1.1e-3--4.8e-3 & 0.00 & 81$\to$7.4 & -1.55 & -1.50 & -1.34 & 0.31--0.77 \\
fp32, b128, const. & DPO & 5e-6 & 7.0e-4--2.1e-3 & 0.00 & 224$\to$77 & -1.00 & -1.69 & -0.97 & 0.88--1.06 \\
fp32, b128, const. & DPO & 5e-5 & 6.7e-3--2.1e-2 & 0.00 & 76$\to$23 & -1.05 & -1.22 & -0.78 & 0.65--0.80 \\
fp32, b128, linear & SFT & 5e-6 & 5.9e-4--3.4e-3 & 0.00 & 167$\to$10 & -1.65 & -1.56 & -1.53 & 0.32--0.93 \\
fp32, b128, linear & DPO & 5e-6 & 2.7e-4--1.2e-3 & 0.00 & 687$\to$125 & -1.13 & -1.26 & -0.80 & 1.02--1.19 \\
fp32, b128, linear & DPO & 5e-5 & 2.4e-3--1.2e-2 & 0.00 & 123$\to$28 & -0.93 & -1.25 & -0.93 & 0.70--0.90 \\
fp32, b8, const., GSM8K & GRPO & 1e-6 & 1.3e-4--3.4e-4 & 0.00 & 283$\to$300 & +0.02 & +0.02 & +0.06 & 1.37--1.50 \\
fp32, b8, const., GSM8K & GRPO & 3e-6 & 4.0e-4--9.5e-4 & 0.00 & 249$\to$294 & +0.25 & +0.71 & +0.23 & 1.37--1.44 \\
fp32, b8, const., GSM8K & GRPO & 1e-5 & 1.3e-3--2.3e-3 & 0.00 & 256$\to$232 & -0.27 & +0.20 & +1.25 & 1.28--1.32 \\
fp32, b8, const., 10k steps & SFT & 5e-6 & 4.3e-3--1.7e-2 & 0.00 & 9.7$\to$3.5 & -0.73 & -0.82 & -0.69 & 0.39--0.42 \\
fp32, b8, const., 10k steps & DPO & 5e-6 & 1.9e-3--1.1e-2 & 0.00 & 118$\to$31 & -0.85 & -0.81 & -0.69 & 0.72--0.89 \\
\bottomrule
\end{tabular}

%% file: tables/tab_r4_repro_family.tex
\begin{tabular}{llllrl}
\toprule
anchor & recipe & $\relDW$ & $\rho_{\text{iso}}$ (mixed) & slope & $c_{\text{sign}}/c_{\text{profile}}$ \\
\midrule
OLMo-2-1B-SFT & released recipe (dpo\_norm, $\beta$ 5, OLMo mix, b128, linear, $2.5\times10^{-6}$) & 2.9e-4--8.6e-4 & 3{,}271$\to$1{,}447 & -0.77 & 1.26--1.32 \\
OLMo-2-1B-SFT & plain: sigmoid, $\beta$ 0.1, UltraFeedback, b8, const., $5\times10^{-6}$ & 4.1e-4--1.4e-3 & 415$\to$205 & -0.52 & 0.96--0.99 \\
OLMo-2-1B-SFT & plain, OLMo mix & 4.2e-4--1.4e-3 & 430$\to$191 & -0.62 & 0.93--1.01 \\
OLMo-2-1B-SFT & plain, dpo\_norm $\beta$ 5 & 4.1e-4--1.2e-3 & 206$\to$282 & +0.25 & 1.00--1.02 \\
Qwen2.5-0.5B & plain & 9.5e-4--3.0e-3 & 185$\to$78 & -0.77 & 0.84--1.01 \\
Qwen2.5-0.5B & plain, dpo\_norm $\beta$ 5 & 9.8e-4--2.7e-3 & 90$\to$132 & +0.42 & 1.00--1.02 \\
Qwen2.5-0.5B & plain, OLMo mix & 9.8e-4--3.0e-3 & 176$\to$83 & -0.63 & 0.83--0.99 \\
Qwen2.5-0.5B-Instr. & plain & 9.7e-4--3.1e-3 & 202$\to$71 & -0.84 & 0.82--0.92 \\
\bottomrule
\end{tabular}

%% file: tables/tab_r4_seeds.tex
\begin{tabular}{lrlll}
\toprule
condition & seeds & ref.\ $\relDW$ & $\rho_{\text{iso}}$ at ref.\ (range) & slope (mean $\pm$ s.d.) \\
\midrule
SFT, bf16 params, b8 const. & 3 & 2.0e-3 & 6.8 (6.7--6.8) & -1.50 $\pm$ 0.01 \\
SFT, fp32, b8 const. & 3 & 2.0e-3 & 32 (31--32) & -1.37 $\pm$ 0.00 \\
DPO, bf16 params, b8 const. & 3 & 5.0e-4 & 70 (69--72) & -0.24 $\pm$ 0.06 \\
DPO, fp32, b8 const. & 3 & 1.0e-3 & 177 (171--184) & -0.64 $\pm$ 0.05 \\
DPO, fp32, b8 linear & 3 & 1.0e-3 & 179 (172--188) & -0.73 $\pm$ 0.02 \\
DPO, fp32, b128 const. & 3 & 1.0e-3 & 165 (161--167) & -1.02 $\pm$ 0.04 \\
DPO, fp32, b128 linear & 3 & 1.0e-3 & 167 (158--173) & -1.21 $\pm$ 0.08 \\
Qwen2.5-0.5B, own trainer, plain & 3 & 1.0e-3 & 151 (132--180) & -0.61 $\pm$ 0.14 \\
Qwen2.5-0.5B, released loss & 3 & 1.0e-3 & 108 (90--127) & +0.18 $\pm$ 0.22 \\
Qwen2.5-0.5B, released data & 3 & 1.0e-3 & 169 (166--175) & -0.66 $\pm$ 0.04 \\
Qwen2.5-0.5B, Instruct anchor & 3 & 1.0e-3 & 165 (143--194) & -0.72 $\pm$ 0.12 \\
OLMo-2-1B-SFT, plain & 3 & 8.4e-4 & 273 (237--311) & -0.41 $\pm$ 0.16 \\
OLMo-2-1B-SFT, released loss & 3 & 8.4e-4 & 263 (256--271) & +0.32 $\pm$ 0.19 \\
OLMo-2-1B-SFT, released data & 3 & 8.4e-4 & 301 (281--315) & -0.75 $\pm$ 0.11 \\
OLMo-2-1B-SFT, released loss + data & 2 & 8.4e-4 & 485 (476--493) & +0.08 $\pm$ 0.04 \\
OLMo-2-1B-SFT, norm. loss, $\beta$ 0.1 & 2 & 8.4e-4 & 734 (614--853) & +0.00 $\pm$ 0.51 \\
OLMo-2-1B-SFT, sigmoid, $\beta$ 5 & 2 & 8.4e-4 & 190 (182--197) & -0.23 $\pm$ 0.03 \\
OLMo-2-1B-SFT, released recipe & 2 & 8.4e-4 & 1{,}490 (1{,}472--1{,}508) & -0.77 $\pm$ 0.01 \\
\bottomrule
\end{tabular}

%% file: tables/tab_r5_loo.tex
\begin{tabular}{lrlllrrr}
\toprule
run (600 steps at the released settings) & seeds & $\relDW$ & $\rho$ at $4\times10^{-4}$: run / released (released $\div$ run) & $\rho$ at $5\times10^{-4}$ & slope & $c_{\text{sign}}/c_{\text{profile}}$ & $\rho_{\text{sign}}$ \\
\midrule
plain loss + plain data & 3 & 5.8e-5--5.1e-4 & 707 (685--737) / 2{,}711 (3.8$\times$) & 511 (502--520) / 2{,}353 (4.6$\times$) & -0.95 $\pm$ 0.01 & 1.01--1.02 & 462--482 \\
released minus its loss & 3 & 5.7e-5--5.2e-4 & 495 (446--549) / 2{,}711 (5.5$\times$) & 323 (311--335) / 2{,}353 (7.3$\times$) & -1.55 $\pm$ 0.04 & 0.92--1.02 & 293--524 \\
released minus its data & 3 & 5.4e-5--5.1e-4 & 1{,}528 (1{,}426--1{,}646) / 2{,}711 (1.8$\times$) & 1{,}282 (1{,}276--1{,}289) / 2{,}353 (1.8$\times$) & -0.27 $\pm$ 0.06 & 1.23--1.25 & 977--1{,}006 \\
released minus its context length & 3 & 5.6e-5--5.6e-4 & 2{,}804 (2{,}722--2{,}870) / 2{,}711 (1.0$\times$) & 2{,}307 (2{,}227--2{,}350) / 2{,}353 (1.0$\times$) & -0.48 $\pm$ 0.02 & 1.29--1.31 & 1{,}472--1{,}552 \\
\bottomrule
\end{tabular}

%% file: tables/tab_r5_lrctl.tex
\begin{tabular}{lrlllrrr}
\toprule
run (released loss and data) & seeds & $\relDW$ & $\rho$ at $4\times10^{-4}$: run / released (released $\div$ run) & $\rho$ at $5\times10^{-4}$ & slope & $c_{\text{sign}}/c_{\text{profile}}$ & $\rho_{\text{sign}}$ \\
\midrule
constant $2.5\times10^{-6}$, batch 128, 600 steps & 3 & 3.0e-4--6.9e-4 & 2{,}236 (2{,}031--2{,}363) / 2{,}711 (1.2$\times$) & 2{,}113 (1{,}944--2{,}321) / 2{,}353 (1.1$\times$) & -0.21 $\pm$ 0.04 & 1.25--1.30 & 1{,}282--1{,}385 \\
constant $2.5\times10^{-6}$, batch 8, 800 steps & 3 & 2.7e-4--7.6e-4 & 500 (363--584) / 2{,}711 (5.4$\times$) & 573 (442--716) / 2{,}353 (4.1$\times$) & +0.08 $\pm$ 0.36 & 1.04--1.11 & 583--704 \\
\bottomrule
\end{tabular}

%% file: tables/tab_r5_7b.tex
\begin{tabular}{llllllrlr}
\toprule
arm & lr & $\relDW$ & $\rho$ at $5\times10^{-4}$ & $10^{-3}$ & $3\times10^{-3}$ & slope & $c_{\text{sign}}/c_{\text{profile}}$ & unchanged \\
\midrule
SFT & 5e-7 & 3.0e-4--1.3e-3 & 804 & 235 & -- & -1.46 & 0.17--1.91 & 0.045 \\
SFT & 5e-6 & 1.9e-3--6.0e-3 & -- & -- & 79 & -1.19 & 0.64--0.66 & 0.045 \\
DPO & 5e-7 & 1.7e-4--5.8e-4 & 647 & -- & -- & -0.26 & 1.57--1.79 & 0.043 \\
DPO & 5e-6 & 1.5e-3--5.5e-3 & -- & -- & 227 & -0.73 & 1.27--1.66 & 0.043 \\
\bottomrule
\end{tabular}

%% file: tables/tab_r5_trunc.tex
\begin{tabular}{llrrrrrr}
\toprule
delta & threshold (ulp) & unchanged & $\relDW$ & $\rho_{\text{iso}}$ & $\rho_{\text{profile}}$ & $\rho_{\text{sign}}$ & $c_{\text{sign}}/c_{\text{profile}}$ \\
\midrule
OLMo-2-1B DPO & as stored & 0.48 & 8.52e-4 & 1{,}449 & 1{,}283 & 1{,}015 & 1.26 \\
 & 1 & 0.48 & 8.51e-4 & 1{,}450 & 1{,}283 & 1{,}015 & 1.26 \\
 & 2 & 0.75 & 6.04e-4 & 1{,}589 & 1{,}367 & 827 & 1.65 \\
 & 4 & 0.88 & 4.17e-4 & 1{,}483 & 1{,}277 & 547 & 2.33 \\
 & 8 & 0.94 & 2.93e-4 & 1{,}192 & 1{,}038 & 327 & 3.17 \\
\addlinespace[2pt]
T\"ulu-3-8B DPO & as stored & 0.76 & 8.40e-4 & 22{,}570 & 21{,}557 & 9{,}844 & 2.19 \\
 & 1 & 0.76 & 8.38e-4 & 22{,}557 & 21{,}539 & 9{,}822 & 2.19 \\
 & 2 & 0.92 & 4.36e-4 & 20{,}386 & 17{,}946 & 3{,}368 & 5.33 \\
 & 4 & 0.96 & 2.81e-4 & 13{,}084 & 11{,}183 & 1{,}113 & 10.05 \\
\addlinespace[2pt]
T\"ulu-3-8B RLVR & as stored & 0.79 & 7.07e-4 & 621 & 593 & 480 & 1.24 \\
 & 1 & 0.79 & 7.05e-4 & 620 & 592 & 478 & 1.24 \\
 & 2 & 0.93 & 3.56e-4 & 326 & 289 & 115 & 2.52 \\
 & 4 & 0.97 & 2.28e-4 & 182 & 157 & 36 & 4.31 \\
\addlinespace[2pt]
Olmo-3-7B Instruct DPO & as stored & 0.67 & 8.84e-4 & 6{,}165 & 5{,}776 & 1{,}625 & 3.55 \\
 & 1 & 0.67 & 8.83e-4 & 6{,}165 & 5{,}777 & 1{,}625 & 3.56 \\
 & 2 & 0.86 & 5.26e-4 & 6{,}770 & 6{,}501 & 881 & 7.38 \\
 & 4 & 0.93 & 3.58e-4 & 5{,}675 & 5{,}528 & 455 & 12.14 \\
\addlinespace[2pt]
\bottomrule
\end{tabular}

%% file: tables/tab_r4_stage1.tex
\begin{tabular}{rrrrrrrrrrrrrrr}
\toprule
 & & \multicolumn{2}{c}{plain-math} & \multicolumn{2}{c}{WikiText} & \multicolumn{4}{c}{mixed, by domain} & \multicolumn{3}{c}{mixed, pooled} & & \\
\cmidrule(lr){3-4}\cmidrule(lr){5-6}\cmidrule(lr){7-10}\cmidrule(lr){11-13}
width (steps) & $\relDW$ & exact & FD & exact & FD & web & chat & math & code & exact & FD & median & $\rho_{\text{sign}}$ & $c_{\text{sign}}/c_{\text{profile}}$ \\
\midrule
814 & 0.027 & 11 & 5.9 & 5.8 & 5.2 & 15 & 3.9 & 750 & 37 & 322 & 4.7 & 5.2 & 616 & 0.54 \\
1{,}814 & 0.037 & 18 & 3.4 & 4.3 & 3.3 & 4.0 & 2.7 & 186 & 21 & 82 & 3.1 & 3.2 & 164 & 0.54 \\
2{,}814 & 0.045 & 3.2 & 2.9 & 2.7 & 2.5 & 3.6 & 2.1 & 118 & 8.5 & 52 & 2.4 & 2.4 & 105 & 0.53 \\
3{,}814 & 0.050 & 4.4 & 2.5 & 2.9 & 2.3 & 3.6 & 1.8 & 140 & 12 & 62 & 2.1 & 2.1 & 127 & 0.53 \\
4{,}814 & 0.055 & 4.4 & 2.0 & 2.5 & 2.0 & 2.9 & 1.6 & 62 & 11 & 29 & 2.0 & 1.9 & 61 & 0.52 \\
5{,}814 & 0.058 & 3.0 & 2.1 & 2.0 & 2.0 & 3.1 & 1.5 & 39 & 5.5 & 18 & 1.8 & 1.6 & 37 & 0.54 \\
7{,}814 & 0.065 & 2.8 & 1.8 & 1.8 & 1.7 & 2.5 & 1.3 & 7.2 & 4.9 & 4.7 & 1.6 & 1.5 & 9.5 & 0.55 \\
10{,}814 & 0.073 & 1.9 & 1.8 & 1.7 & 1.6 & 1.6 & 1.2 & 10 & 4.9 & 5.9 & 1.5 & 1.3 & 12 & 0.56 \\
30{,}814 & 0.102 & 1.2 & 1.1 & 1.5 & 1.1 & 1.3 & 0.77 & 25 & 3.8 & 12 & 1.1 & 0.92 & 19 & 0.69 \\
100{,}814 & 0.156 & 1.3 & 1.3 & 0.9 & 1.1 & 0.74 & 0.64 & 2.4 & 1.4 & 1.5 & 1.0 & 0.83 & 1.2 & 1.34 \\
200{,}814 & 0.220 & 1.2 & 1.1 & 1.1 & 1.4 & 0.77 & 0.63 & 13 & 1.9 & 6.0 & 1.2 & 0.89 & 4.2 & 1.51 \\
400{,}814 & 0.356 & 0.99 & 1.1 & 0.95 & 1.2 & 0.65 & 0.62 & 0.6 & 1.6 & 0.81 & 1.0 & 0.73 & 0.76 & 1.26 \\
700{,}814 & 0.600 & 0.86 & 0.91 & 1.3 & 1.2 & 0.72 & 0.74 & 1.3 & 2.5 & 1.3 & 1.1 & 0.87 & 1.6 & 1.13 \\
1{,}000{,}814 & 0.918 & 1.1 & 1.3 & 1.2 & 1.2 & 0.89 & 1.0 & 3.0 & 2.6 & 2.1 & 1.1 & 1.0 & 2.8 & 1.09 \\
\bottomrule
\end{tabular}

%% file: tables/tab_windows_olmo.tex
{\footnotesize\setlength{\tabcolsep}{3pt}
\resizebox{\linewidth}{!}{%
\begin{tabular}{rrl rrr rrr rrr r}
\toprule
 & & & \multicolumn{3}{c}{plain-math} & \multicolumn{3}{c}{WikiText} & \multicolumn{3}{c}{mixed} & mixed, median \\
\cmidrule(lr){4-6}\cmidrule(lr){7-9}\cmidrule(lr){10-12}
width (steps) & rel$\Delta W$ & read & iso & prof. & sign & iso & prof. & sign & iso & prof. & sign & iso \\
\midrule
814 & 0.027 & $\alpha=0.3$ & 5.94 & 6.06 & 10.5 & 5.17 & 5.52 & 8.50 & 4.65 & 4.96 & 7.92 & 4.01 \\
1,814 & 0.037 & $\alpha=0.3$ & 3.41 & 3.58 & 6.28 & 3.26 & 3.58 & 5.67 & 3.09 & 3.36 & 5.54 & 2.58 \\
2,814 & 0.045 & $\alpha=0.3$ & 2.94 & 3.14 & 5.51 & 2.46 & 2.73 & 4.55 & 2.41 & 2.65 & 4.42 & 2.02 \\
3,814 & 0.050 & $\alpha=0.3$ & 2.54 & 2.75 & 4.99 & 2.28 & 2.54 & 4.25 & 2.14 & 2.36 & 3.99 & 1.75 \\
4,814 & 0.055 & $\alpha=0.3$ & 2.04 & 2.23 & 3.79 & 2.04 & 2.27 & 3.81 & 2.00 & 2.22 & 3.80 & 1.58 \\
5,814 & 0.058 & $\alpha=0.3$ & 2.12 & 2.33 & 4.36 & 2.04 & 2.28 & 3.72 & 1.82 & 2.02 & 3.45 & 1.50 \\
7,814 & 0.065 & $\alpha=0.3$ & 1.75 & 1.96 & 3.60 & 1.69 & 1.90 & 3.26 & 1.63 & 1.81 & 3.05 & 1.30 \\
10,814 & 0.073 & certified & 1.81 & 2.05 & 3.55 & 1.58 & 1.78 & 2.86 & 1.45 & 1.62 & 2.73 & 1.17 \\
30,814 & 0.102 & certified & 1.14 & 1.35 & 1.80 & 1.08 & 1.22 & 1.92 & 1.08 & 1.21 & 1.63 & 0.85 \\
100,814 & 0.156 & certified & 1.27 & 1.37 & 1.21 & 1.11 & 1.19 & 1.36 & 1.00 & 1.04 & 0.74 & 0.93 \\
200,814 & 0.220 & certified & 1.08 & 1.23 & 1.06 & 1.41 & 1.50 & 1.46 & 1.24 & 1.32 & 0.86 & 1.04 \\
400,814 & 0.356 & certified & 1.09 & 1.47 & 1.23 & 1.17 & 1.41 & 1.43 & 1.01 & 1.25 & 1.04 & 0.95 \\
700,814 & 0.600 & certified & 0.91 & 1.26 & 1.19 & 1.17 & 1.58 & 1.59 & 1.09 & 1.45 & 1.73 & 0.90 \\
1,000,814 & 0.918 & certified & 1.28 & 1.93 & 2.46 & 1.19 & 1.69 & 1.87 & 1.14 & 1.69 & 1.87 & 1.07 \\
\bottomrule
\end{tabular}}}

%% file: tables/tab_r5_pythia.tex
\begin{tabular}{rrrrrrrrrrr}
\toprule
 & & \multicolumn{2}{c}{plain-math} & \multicolumn{2}{c}{WikiText} & \multicolumn{2}{c}{mixed} & & & \\
\cmidrule(lr){3-4}\cmidrule(lr){5-6}\cmidrule(lr){7-8}
width (steps) & $\relDW$ & exact & FD & exact & FD & exact & FD & $\rho_{\text{sign}}$ & $c_{\text{sign}}/c_{\text{profile}}$ & top-1\% share \\
\midrule
1{,}000 & 0.012 & 4.0 & 4.5 & 5.4 & 6.3 & 5.0 & 5.7 & 14 & 0.37 & 0.11 \\
2{,}000 & 0.017 & 2.7 & 3.2 & 3.2 & 3.7 & 3.0 & 3.5 & 8.7 & 0.35 & 0.11 \\
4{,}000 & 0.023 & 1.8 & 2.1 & 1.9 & 2.2 & 1.9 & 2.2 & 5.5 & 0.32 & 0.11 \\
8{,}000 & 0.032 & 0.94 & 1.2 & 1.1 & 1.4 & 1.1 & 1.4 & 3.2 & 0.29 & 0.11 \\
16{,}000 & 0.047 & 0.54 & 0.79 & 0.62 & 0.84 & 0.58 & 0.87 & 1.6 & 0.27 & 0.11 \\
43{,}000 & 0.116 & 0.46 & 0.74 & 0.46 & 0.81 & 0.65 & 1.2 & 1.6 & 0.26 & 0.16 \\
\bottomrule
\end{tabular}

%% file: tables/tab_windows_pythia.tex
{\footnotesize\setlength{\tabcolsep}{3pt}
\resizebox{\linewidth}{!}{%
\begin{tabular}{rrr rrr rrr rrr rr}
\toprule
 & \multicolumn{2}{c}{rel$\Delta W$} & \multicolumn{3}{c}{plain-math} & \multicolumn{3}{c}{WikiText} & \multicolumn{3}{c}{mixed} & \multicolumn{2}{c}{end, mixed} \\
\cmidrule(lr){2-3}\cmidrule(lr){4-6}\cmidrule(lr){7-9}\cmidrule(lr){10-12}\cmidrule(lr){13-14}
width & end & start & end & start & ratio & end & start & ratio & end & start & ratio & prof. & sign \\
\midrule
1,000 & 0.012 & 0.027 & 4.54 & 2.94 & 1.55 & 6.30 & 3.51 & 1.80 & 5.74 & 3.16 & 1.82 & 5.82 & 15.3 \\
2,000 & 0.017 & 0.035 & 3.20 & 1.94 & 1.65 & 3.75 & 2.24 & 1.68 & 3.54 & 2.03 & 1.75 & 3.51 & 9.83 \\
4,000 & 0.023 & 0.045 & 2.10 & 1.15 & 1.82 & 2.19 & 1.38 & 1.59 & 2.23 & 1.26 & 1.77 & 2.11 & 6.31 \\
8,000 & 0.032 & 0.058 & 1.19 & 0.76 & 1.57 & 1.45 & 0.84 & 1.72 & 1.42 & 0.75 & 1.90 & 1.23 & 3.89 \\
16,000 & 0.047 & 0.073 & 0.79 & 0.43 & 1.84 & 0.84 & 0.51 & 1.65 & 0.87 & 0.44 & 1.97 & 0.65 & 2.23 \\
43,000 & 0.116 & 0.119 & 0.74 & 0.20 & 3.68 & 0.81 & 0.23 & 3.49 & 1.19 & 0.21 & 5.62 & 0.76 & 2.46 \\
\bottomrule
\end{tabular}}}

%% file: tables/tab_long.tex
{\small\setlength{\tabcolsep}{5pt}
\begin{tabular}{llrrrr}
\toprule
 & & \multicolumn{2}{c}{plain-math} & \multicolumn{2}{c}{WikiText} \\
\cmidrule(lr){3-4}\cmidrule(lr){5-6}
 & $\relDW$ bin & short runs & long runs & short runs & long runs \\
\midrule
SFT & $[1e-04,3e-04)$ & 82.1 (9) & 65.3 (6) & 34.2 (1) & 22.3 (6) \\
 & $[3e-04,1e-03)$ & 44.1 (14) & 48.5 (34) & 17.3 (14) & 17.0 (34) \\
 & $[1e-03,3e-03)$ & 12.6 (12) & 11.0 (1) & 5.70 (12) & 4.73 (1) \\
 & $[3e-03,1e-02)$ & 4.71 (34) & 3.97 (39) & 2.58 (34) & 1.40 (39) \\
 & $[1e-02,3e-02)$ & 6.11 (22) & -- & 7.67 (22) & -- \\
\addlinespace[2pt]
DPO & $[1e-04,3e-04)$ & 67.1 (4) & -- & 46.4 (4) & -- \\
 & $[3e-04,1e-03)$ & 92.2 (25) & 104 (7) & 44.0 (25) & 48.2 (7) \\
 & $[1e-03,3e-03)$ & 80.6 (18) & 57.8 (33) & 42.7 (18) & 34.6 (33) \\
 & $[3e-03,1e-02)$ & 94.8 (7) & -- & 39.0 (7) & -- \\
 & $[1e-02,3e-02)$ & 43.3 (18) & 18.6 (8) & 24.3 (18) & 15.9 (8) \\
 & $[3e-02,1e-01)$ & -- & 10.4 (32) & -- & 7.99 (32) \\
\addlinespace[2pt]
GRPO & $[3e-05,1e-04)$ & 143 (1) & 112 (6) & -- & 14.9 (1) \\
 & $[1e-04,3e-04)$ & 463 (3) & 402 (31) & -- & 22.0 (31) \\
 & $[3e-04,1e-03)$ & 690 (6) & 642 (9) & 40.2 (6) & 37.3 (9) \\
\addlinespace[2pt]
\bottomrule
\end{tabular}}

%% file: tables/tab_extrap.tex
{\footnotesize\setlength{\tabcolsep}{3pt}
\resizebox{\linewidth}{!}{%
\begin{tabular}{lllll l rrll l r r}
\toprule
 & & & & \multicolumn{2}{c}{GSM8K, 200 problems} & \multicolumn{4}{c}{MATH-500, 200 problems} & & & \\
\cmidrule(lr){5-6}\cmidrule(lr){7-10}
delta & stage & rel$\Delta W$ & prediction & acc 0/1/1.5 & verdict & acc 0/1/1.5/2 & kept 1.5 & verdict 1.5 & verdict 2.0 & pooled & stop 1.0/1.5 & $\alpha^*$ \\
\midrule
\texttt{olmo2\_7b\_rlvr} & RLVR & $3.27\times10^{-4}$ & \textsc{tolerant} & 0.75/0.76/0.74 (new) & unscorable & 0.31/0.37/0.36/0.38 & -- & unscorable & unscorable & unscorable & 0.92/0.91 & 1.00 \\
\texttt{tulu\_rlvr} & RLVR & $7.07\times10^{-4}$ & \textsc{tolerant} & 0.83/0.82/0.82 (earlier) & unscorable & 0.41/0.38/0.41/0.38 & -- & unscorable & unscorable & unscorable & 0.86/0.78 & 1.00 \\
\texttt{olmo\_rlvr\_think} & RLVR (think) & $8.06\times10^{-4}$ & \textsc{tolerant} & 0.87/0.87/0.85 (earlier) & unscorable & 0.62/0.62/0.61/0.62 & -- & unscorable & unscorable & unscorable & 0.58/0.57 & 1.00 \\
\texttt{deepscaler\_rlvr} & RLVR (GRPO) & $8.10\times10^{-4}$ & \textsc{tolerant} & 0.74/0.69/0.55 (earlier) & unscorable & 0.32/0.29/0.24/0.14 & -- & unscorable & unscorable & unscorable & 0.56/0.49 & 1.00 \\
\texttt{olmo\_dpo\_instruct} & DPO & $8.84\times10^{-4}$ & \textsc{tolerant} & 0.76/0.71/0.63 (earlier) & unscorable & 0.41/0.39/0.45/0.45 & -- & unscorable & unscorable & unscorable & 0.92/0.81 & 1.00 \\
\texttt{olmo2\_7b\_dpo} & DPO & $8.85\times10^{-4}$ & \textsc{tolerant} & 0.61/0.74/0.74 (new) & hit & 0.16/0.31/0.33/0.28 & 1.10 & hit & hit & hit & 0.90/0.88 & 1.00 \\
\texttt{olmo\_rlvr\_instruct} & RLVR & $1.08\times10^{-3}$ & \textsc{tolerant} & 0.71/0.70/0.68 (earlier) & unscorable & 0.36/0.32/0.32/0.29 & -- & unscorable & unscorable & unscorable & 0.90/0.91 & 1.00 \\
\texttt{olmo3\_rlzero\_if} & RLVR from base (IF) & $1.38\times10^{-3}$ & \textsc{tolerant} & 0.73/0.72/0.74 (new) & unscorable & 0.44/0.46/0.46/0.45 & -- & unscorable & unscorable & unscorable & 0.94/0.95 & 1.00 \\
\texttt{zephyr\_dpo} & DPO & $1.40\times10^{-3}$ & \textsc{tolerant} & 0.37/0.37/0.23 (earlier) & unscorable & 0.07/0.04/0.04/0.01 & -- & unscorable & unscorable & unscorable & 0.92/0.89 & 1.00 \\
\texttt{olmo3\_rlzero\_code} & RLVR from base (code) & $1.51\times10^{-3}$ & \textsc{tolerant} & 0.73/0.72/0.75 (new) & unscorable & 0.44/0.47/0.46/0.43 & -- & unscorable & unscorable & unscorable & 0.90/0.92 & 1.00 \\
\texttt{tulu31\_rlvr} & RLVR (GRPO) & $1.96\times10^{-3}$ & \textsc{tolerant} & 0.82/0.89/0.79 (new) & unscorable & 0.41/0.51/0.42/0.37 & 0.15 & miss & miss & miss & 0.81/0.64 & 1.00 \\
\texttt{prorl\_rlvr} & RLVR (ProRL) & $3.20\times10^{-3}$ & \textsc{tolerant} & 0.74/0.61/0.33 (earlier) & unscorable & 0.32/0.27/0.10/0.05 & -- & unscorable & unscorable & unscorable & 0.32/0.10 & 0.30 \\
\texttt{mistral\_sft} & SFT & $8.69\times10^{-3}$ & \textsc{tolerant} & 0.32/0.34/0.02 (new) & unscorable & 0.10/0.09/0.02/0.00 & -- & unscorable & unscorable & unscorable & 0.76/0.55 & 0.30 \\
\texttt{qwen3b\_instruct} & SFT$+$RLHF (instruct) & $1.27\times10^{-2}$ & \textsc{fragile} & 0.76/0.81/0.74 (new) & unscorable & 0.36/0.53/0.56/0.51 & 1.15 & miss & miss & miss & 0.47/0.12 & 0.30 \\
\texttt{qwen7b\_instruct} & SFT$+$RLHF (instruct) & $1.34\times10^{-2}$ & \textsc{fragile} & 0.83/0.88/0.87 (new) & unscorable & 0.49/0.45/0.68/0.68 & -- & unscorable & unscorable & unscorable & 0.53/0.68 & 1.00 \\
\texttt{llama31\_tulu\_sft} & SFT & $2.57\times10^{-2}$ & \textsc{fragile} & 0.47/0.73/0.53 (new) & hit & 0.14/0.20/0.20/0.01 & -- & unscorable & unscorable & hit & 0.86/0.73 & 0.30 \\
\texttt{olmo2\_7b\_sft} & SFT & $3.89\times10^{-2}$ & \textsc{fragile} & 0.66/0.61/0.49 (new) & unscorable & 0.20/0.15/0.05/0.01 & -- & unscorable & unscorable & unscorable & 0.76/0.65 & 1.00 \\
\texttt{qwen05\_instruct} & SFT$+$RLHF (instruct) & $4.18\times10^{-2}$ & \textsc{fragile} & 0.35/0.36/0.28 (new) & unscorable & 0.15/0.30/0.24/0.14 & 0.61 & miss & hit & miss & 0.69/0.18 & 1.00 \\
\texttt{smollm2\_instruct} & SFT$+$DPO (instruct) & $6.34\times10^{-2}$ & \textsc{fragile} & 0.24/0.45/0.04 (new) & hit & 0.05/0.19/0.15/0.01 & 0.71 & miss & hit & miss & 0.22/0.01 & 1.00 \\
\texttt{r1distill\_sft} & SFT (R1 distillation) & $1.19\times10^{-1}$ & \textsc{fragile} & 0.72/0.08/0.00 (earlier) & unscorable & 0.49/0.08/0.00/0.00 & -- & unscorable & unscorable & unscorable & 0.18/0.00 & 0.10 \\
\texttt{qwenmath\_instruct} & SFT$+$RLHF (math instruct) & $1.57\times10^{-1}$ & \textsc{fragile} & 0.72/0.71/0.31 (earlier) & unscorable & 0.49/0.49/0.30/0.00 & -- & unscorable & unscorable & unscorable & 0.81/0.04 & 0.30 \\
\bottomrule
\end{tabular}}}